\documentclass{bmvc2k}

\usepackage{amsfonts}
\usepackage{mathtools}
\usepackage{amsthm}

\usepackage{placeins}

\usepackage{booktabs}

\usepackage{multirow}
\usepackage{makecell}

\usepackage{listings}
\usepackage{algorithm}
\usepackage{algorithmic}

\usepackage{amsmath,amsfonts,bm}

\def\eqref#1{equation~\ref{#1}}

\def\1{\bm{1}}

\DeclareMathAlphabet{\mathsfit}{\encodingdefault}{\sfdefault}{m}{sl}
\SetMathAlphabet{\mathsfit}{bold}{\encodingdefault}{\sfdefault}{bx}{n}

\def\sY{{\mathbb{Y}}}

\DeclareMathOperator*{\argmax}{arg\,max}
\DeclareMathOperator*{\argmin}{arg\,min}

\title{eXplaining to Learn (eX2L): \\ Regularization Using Contrastive Visual Explanation Pairs for Distribution Shifts}

\addauthor{Paulo Mario P. Medina}{pmmedina@seameo-innotech-ecair.org}{1}
\addauthor{Jose Marie Antonio Miñoza}{jmaminoza@seameo-innotech-ecair.org}{1}
\addauthor{Sebastian C. Ibañez}{sibanez@seameo-innotech-ecair.org}{1}

\addinstitution{
 Education Center for AI Research (ECAIR)\\
 Department of Education\\
 Makati, Philippines
}

\runninghead{Medina, Miñoza, Ibañez}{eXplaining to Learn (eX2L)}

\begin{document}

\maketitle

\begin{abstract}
    Despite extensive research into mitigating distribution shifts, many existing algorithms yield inconsistent performance, often failing to outperform baseline Empirical Risk Minimization (ERM) across diverse scenarios and necessitating newer algorithms which can handle scenarios where existing algorithms currently underperform. Furthermore, high algorithmic complexity frequently limits interpretability and offers only an indirect means of addressing spurious correlations.
    We propose eXplaining to Learn (eX2L): an interpretable, explanation-based framework that decorrelates confounding features from a classifier's latent representations during training.
    eX2L achieves this by penalizing the similarity between Grad-CAM activation maps generated by a primary label classifier and those from a concurrently trained confounder classifier.
    On the rigorous Spawrious Many-to-Many Hard Challenge synthetic data benchmark, eX2L achieves an average accuracy (AA) of $82.24\% \pm 3.87\%$ and a worst-group accuracy (WGA) of $66.31\% \pm 8.73\%$, outperforming the current state-of-the-art (SOTA) by $5.49\%$ and $10.90\%$, respectively.
    Beyond its competitive performance, eX2L demonstrates that functional domain invariance can be enforced by explicitly decoupling label and nuisance attributes at the group level.
\end{abstract}

\section{Introduction}
\label{sec:intro}

\begin{figure*}[t]
\centering
\setlength{\tabcolsep}{2pt}
\begin{tabular}{cccccc}
 & \textbf{Original} & \textbf{ERM} & \textbf{GroupDRO} & \textbf{eX2L ($\sY$)} & \textbf{eX2L ($C$)} \\
$\mathbf{y = {<5}}$ & \includegraphics[width=14mm]{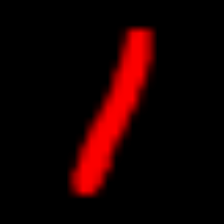} &   \includegraphics[width=14mm]{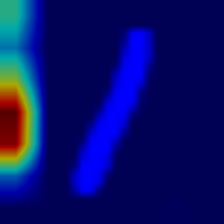} & \includegraphics[width=14mm]{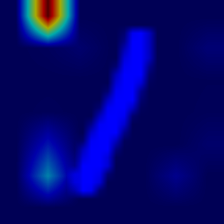} &  \includegraphics[width=14mm]{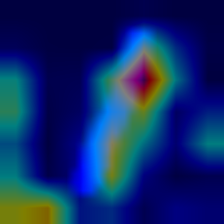} & \includegraphics[width=14mm]{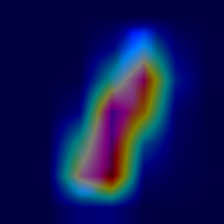} \\
 & \textit{CMNIST} & $\hat{y} = \text{$\geq 5$}$ & $\hat{y} = \text{$\geq 5$}$ & $\hat{y} = \text{$<5$}$ & $c = \text{Red}$ \\[3pt]
$\textbf{y = \textbf{Waterbird}}$ & \includegraphics[width=14mm]{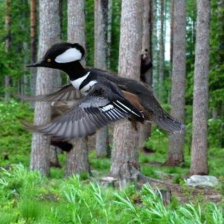} &   \includegraphics[width=14mm]{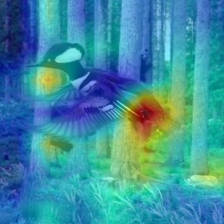} & \includegraphics[width=14mm]{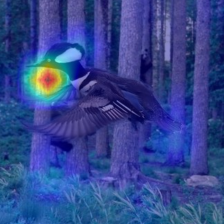} &  \includegraphics[width=14mm]{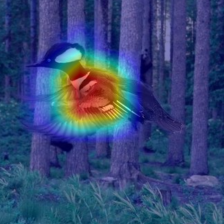} & \includegraphics[width=14mm]{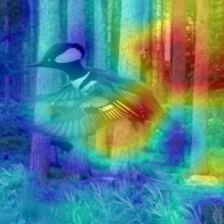} \\
 & \textit{Waterbirds} & $\hat{y} = \text{Landbird}$ & $\hat{y} = \text{Landbird}$ & $\hat{y} = \text{Waterbird}$ & $c = \text{Land}$ \\[3pt]
 $\textbf{y = \textbf{Blonde}}$ & \includegraphics[width=14mm]{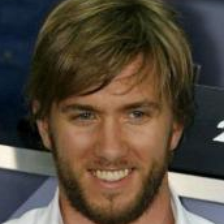} &   \includegraphics[width=14mm]{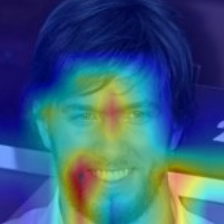} & \includegraphics[width=14mm]{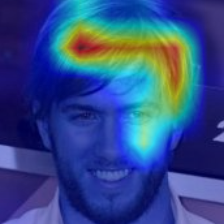} &  \includegraphics[width=14mm]{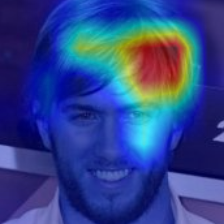} & \includegraphics[width=14mm]{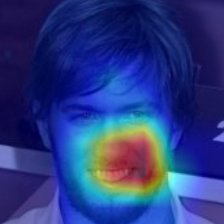} \\
 & \textit{CelebA} & $\hat{y} = \text{Non-blonde}$ & $\hat{y} = \text{Blonde}$ & $\hat{y} = \text{Blonde}$ & $c = \text{Male}$ \\[3pt]
 $\textbf{y = \textbf{Corgi}}$ & \includegraphics[width=14mm]{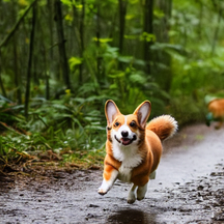} &   \includegraphics[width=14mm]{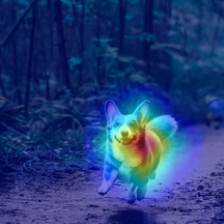} & \includegraphics[width=14mm]{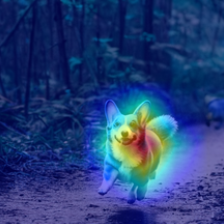} &  \includegraphics[width=14mm]{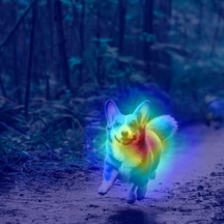} & \includegraphics[width=14mm]{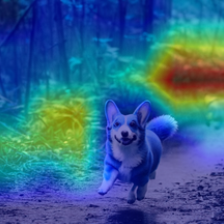} \\
 & \textit{S. O2O Easy} & $\hat{y} = \text{Corgi}$ & $\hat{y} = \text{Corgi}$ & $\hat{y} = \text{Corgi}$ & $c = \text{Jungle}$ \\[3pt]
  $\textbf{y = \textbf{Dachshund}}$ & \includegraphics[width=14mm]{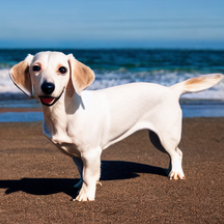} &   \includegraphics[width=14mm]{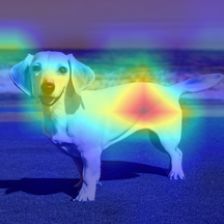} & \includegraphics[width=14mm]{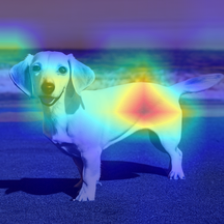} &  \includegraphics[width=14mm]{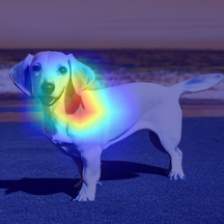} & \includegraphics[width=14mm]{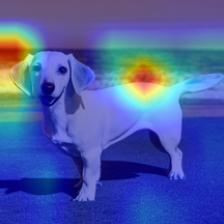} \\
 & \textit{S. M2M Hard} & $\hat{y} = \text{Labrador}$ & $\hat{y} = \text{Labrador}$ & $\hat{y} = \text{Dachshund}$ & $c = \text{Beach}$ \\[3pt]
\end{tabular}
\caption{Qualitative evaluation of feature attribution via Grad-CAM across subpopulation shifts and domain generalization benchmarks. Columns display the true label ($y$), predicted label ($\hat{y}$), true confounder ($c$), and the distinct target layers for eX2L's label classifier ($\sY$) and confounder classifier ($C$). By treating the visual explanation as a static spatial prior, eX2L restricts the primary model's focus strictly to invariant, class-predictive features (the dog's anatomy) while relegating the background and other spurious correlation features to the confounder model, directly accounting for eX2L's 11.91\% WGA improvement over GroupDRO.}
\label{figure:1}
\end{figure*}

Standard machine learning models rely on the assumption that training and test data are independently and identically distributed (i.i.d.) \citep{ye2021towards}. When this assumption is violated, models often fail to generalize, instead relying on spurious correlations, statistical patterns in the training data that are not conceptually predictive of the actual labels \citep{monga2025mitigating, suhail2025shortcut, Koh2020WILDSAB}. These violations typically manifest as two types of distribution shifts: subpopulation shifts and domain generalization.

\FloatBarrier

Subpopulation shifts occur when the proportions of demographic or contextual variables change between training and deployment \citep{Koh2020WILDSAB, yang2023change}. Consider a medical diagnostic tool trained on radiographs: if the training set is $80\%$ White and $20\%$ other ethnicities, but the model is deployed in an Asian country where the population is $90\%$ Asian and $10\%$ White/other, the model may suffer from catastrophic performance degradation due to its initial bias toward White physiological features.

In contrast, domain generalization involves learning from one or more "source domains" to perform on entirely unseen "target domains" at deployment \citep{Koh2020WILDSAB, ye2021towards}. For instance, a model trained on Malaysian wildlife images may struggle to classify animals in Philippine wildlife images due to differing camera qualities, lighting, or background vegetation. In these cases, the model often "cheats" by learning the background or other artifacts rather than the animal itself.

While various loss-level regularization methods exist, they face three primary hurdles:

\begin{enumerate}
  \item \textbf{Blind Regularization}: Many black-box algorithms do not explicitly target the spurious correlations causing the performance drop \citep{wiles2022a}.

  \item \textbf{Inconsistent Performance}: No single algorithm consistently outperforms Empirical Risk Minimization (ERM) across all datasets, reflecting the "no-free-lunch" theorem in machine learning \citep{wiles2022a, gulrajani2021in}.

  \item \textbf{Lack of Interpretability}: Technically sophisticated methods, such as Domain Adversarial Neural Networks (DANN) \citep{ganin2016domain}, rely on mechanisms like gradient reversal that are difficult to explain to non-technical stakeholders, hindering trust in sensitive deployments.
\end{enumerate}

We therefore ask: \textit{can penalizing the spatial overlap of gradient-based explanations from a label and confounder classifier serve as a direct, interpretable regularizer for distribution shift robustness, without sacrificing predictive accuracy?}

To bridge these gaps, we introduce eXplaining to Learn (eX2L), a framework at the intersection of out-of-distribution (OOD) generalization and explainable AI (XAI). eX2L regularizes training by penalizing the spatial overlap between the Grad-CAM explanations of a label classifier and a concurrently trained confounder model. By forcing the primary model to diverge from the confounder's visual focus, eX2L explicitly decouples spurious correlations and promotes domain invariance. Critically, our results demonstrate that interpretability can act as a \textit{catalyst} for predictive robustness rather than a trade-off against it, challenging the prevailing assumption that more interpretable models are necessarily less accurate.

The code implementation for eX2l is publicly available on GitHub\footnote{\url{https://github.com/ppmedina11/eX2L}}.

\section{Related Works}
\paragraph{Subpopulation Shifts and Spurious Correlations} While domain generalization dominates the literature, subpopulation shift has emerged as a distinct challenge where models rely on spurious correlations, non-predictive features that correlate with labels in specific groups \citep{Koh2020WILDSAB, yang2023change}. Current state-of-the-art (SOTA) loss-level regularization, such as GroupDRO, employs minimax fairness to minimize worst-group risk \citep{Sagawa*2020Distributionally}. However, optimizing for the worst-case group often comes at the expense of overall average performance and only indirectly addresses the underlying spurious attributes \citep{shen2025boosting}.

Other approaches which also require group-level annotations like Just Train Twice (JTT) and Deep Feature Reweighting (DFR) address these shifts via data-centric two-step pipelines involving initial training and subsequent fine-tuning on curated subsets \citep{liu2021justtraintwiceimproving, kirichenko2023layerretrainingsufficientrobustness}. Given the fundamental difference in methodology, this study focuses on comparing eX2L against loss-based regularization methods, which offer more streamlined, single-stage training.

\paragraph{Domain Generalization} Standard DG approaches prioritize domain invariance, attempting to minimize the divergence between the empirical loss of available domains and the theoretical loss of the global distribution \citep{liu2023outofdistributiongeneralizationsurvey, ye2021towards}. Despite these theoretical foundations, many DG algorithms fail to consistently outperform Empirical Risk Minimization (ERM) on most image datasets \citep{gulrajani2021in}.

Recent perspectives suggest decomposing the label space into label and nuisance attributes \citep{wiles2022a}. Under this framework, distribution shift is viewed as a problematic correlation between labels and nuisances in the source domain. eX2L builds on this motivation by explicitly decoupling these attributes through explanation-based learning.

\paragraph{Explanation-based Learning} Existing explanation-based learning utilizes adversarial or contrastive training to align robust and non-robust representations \citep{han-tsvetkov-2021-influence-tuning, dammu2023detecting}. For example, \citet{hagos2022identifyingspuriouscorrelationscorrecting} used Grad-CAM maps alongside expensive ground-truth masks to reduce spurious reliance, while \citet{dammu2023detecting} used several local surrogate models per class and compared their Grad-CAM maps with the primary classifier's Grad-CAM maps, only to detect spurious visual attributes. However, eX2L departs from these works in two significant ways:

\begin{enumerate}
    \item It requires only two concurrent classifiers (label and confounder), and it does not require ground-truth annotation masks, only confounder labels.
    \item It utilizes Grad-CAM pairs not just for detection, but to automate the debiasing process, directly improving downstream performance without requiring manual annotation masks.
\end{enumerate}

\section{Proposed Method}

\subsection{Problem Setting}
We define a training set of images $\mathcal{X}$, labels $\mathcal{Y}$, and \textit{a priori} confounding variables $C$. Following the group-based framework of \citet{Sagawa*2020Distributionally}, each input $\mathbf{x} \in \mathcal{X}$ belongs to a group $g$ defined by the Cartesian product of labels and confounders: $g \in C \times \mathcal{Y}$.

For the primary classification task, we define a latent representation function $h_{y}: \mathcal{X} \rightarrow Z$ and a classifier $f_{y}: Z \rightarrow \mathcal{Y}$. Following \citet{ming2021impactspuriouscorrelationoutofdistribution}, we assume that in the presence of distribution shifts the latent representation $Z$ admits the decomposition:
\begin{equation}
    Z = Z_{\text{inv}} + Z_{\text{spu}}
\end{equation}
where $Z_{\text{inv}}$ represents invariant features (conceptually predictive of labels) and $Z_{\text{spu}}$ represents spurious/confounding features (predictive of the group $g$ but not the label $\mathcal{Y}$). Because $P(Z_{\text{spu}} | \mathcal{X}_{\text{train}})$ differs significantly from $P(Z_{\text{spu}} | \mathcal{X}_{\text{test}})$, the model's reliance on $Z_{\text{spu}}$ leads to performance degradation at deployment \citep{ming2021impactspuriouscorrelationoutofdistribution}.

To isolate and remove $Z_{\text{spu}}$, we introduce a secondary confounder model $(h_{c}, f_{c})$ trained to predict the confounder labels $C$. While a direct penalty on the similarity of representations $Z$ and $Z_{c}$ might seem intuitive, such an approach is not spatially target-specific. It may penalize relevant features rather than specific confounding regions.

\subsection{The eX2L Algorithm}
To ensure the penalty is target-specific, we utilize Grad-CAM \citep{Selvaraju_2019}, chosen over input-gradient methods (e.g., SHAP, Integrated Gradients) for its faithfulness to model reparametrization rather than input structure \citep{adebayo2018sanity} (see Appendix~\ref{appendix:b}). For a target class $t$, the importance weight $\alpha^{t}_{k}$ for the $k$th activation map $A^k$ is acquired through a global average pooling of the target logit $\hat{t}$'s gradients with respect to each map's pixel $A^{k}_{i, j}$:

\begin{equation}
    \alpha^{t}_{k} = \frac{1}{P} \sum_{i}\sum_{j} \frac{\partial{\hat{t}^t}}{\partial{A^{k}_{i, j}}}
\end{equation}

where $P$ is the total number of pixels in the map.

The heatmap, $L^{t}_{\text{Grad-CAM}} = \text{ReLU} \left(\sum_{k} \alpha^{t}_{k} A^{k}\right) $, highlights the regions the model prioritizes \citep{Selvaraju_2019}, allowing eX2L to minimize the spatial overlap between the label and confounder model explanations. The gradients used to compute the heatmap are detached from the computational graph, while the final convolutional layer's activation maps used for the heatmap remain differentiable. This selective detachment avoids the second-order computational cost of backpropagating through the gradient computation itself. Moreover, to avoid high gradient magnitudes, the heatmaps are normalized using min-max scaling on a per-sample basis ($\text{MinMax}_{\text{sample}}$).

If the label model and the confounder model rely on the same pixels, the label model is likely operating by \textit{shortcut learning} using spurious background cues. Forcing their heat maps to be spatially disjoint compels the label model to identify its own unique, invariant features.
Using the heatmaps for the label and confounder, we define the similarity loss $l_{\text{sim}}$ as:
\begin{equation}
    \mathcal{L}_{\text{sim}} = S(L^{y}_{\text{Grad-CAM}}, L^{c}_{\text{Grad-CAM}})
\end{equation}
where $S$ is a similarity metric (e.g., Cosine Similarity or Negative Mean Squared Error). The total eX2L training objective $\hat{R}_{\text{eX2L}}$ is formulated as:
\begin{equation}
    \begin{split}
    \hat{R}_{\text{eX2L}}(\theta, \phi) = \mathbb{E}_{(\mathbf{x}, y, c) \sim D_{\text{train}}} [\mathcal{L}(\mathbf{x}, y; \theta) + \\
    \lambda_c \mathcal{L}(\mathbf{x}, c; \phi) + \lambda_{\text{sim}} \mathcal{L}_{\text{sim}}(\mathbf{x}, y, c; \theta, \phi)]
    \end{split}
\end{equation}

where $\theta$ refers to the label model's parameters, $\phi$ refers to the confounder model's parameters, $\lambda_c$ refers to the confounder model hyperparameter, and the $\lambda_{\text{sim}}$ refers to the similarity loss hyperparameter.

\subsection{Optimization and Sampling}
We define \textit{functional domain invariance} as the property of a learned representation whose distribution is invariant across environments, even when the model is trained without environment labels or adversarial objectives. As shown in Section~\ref{sec:results}, eX2L enforces this through the Grad-CAM similarity penalty.

The objective is to find the parameters $\theta^*$ that maximize the mean of the Worst-Group Accuracy (WGA) and Average Accuracy (AA) at validation time:
\begin{equation}
    WGA = \min_{g} \mathbb{E}_{(\mathbf{x}, y) \sim D_{\text{test}}^{g}} [\1_{\hat{y}_{\theta} = y}]
\end{equation}

\begin{equation}
    AA = \mathbb{E}_{(\mathbf{x}, y) \sim D_{\text{test}}} [\1_{\hat{y}_{\theta} = y}]
\end{equation}

\begin{equation}
    \theta^{*} = \argmax_{\theta \in \Theta} \frac{WGA(\theta) + AA(\theta)}{2}
\end{equation}

Note that $\theta^{*}$ defines the \textit{model selection criterion} used for checkpointing, not the training objective itself; the model is trained via $\hat{R}_{\text{eX2L}}$, while $\theta^{*}$ selects the checkpoint that best balances WGA and AA. To optimize this, we contrast two sampling strategies:
\begin{itemize}
    \item \textbf{Random Sampling:} Follows the prevailing training distribution.
    \item \textbf{Uniform Group Sampling:} Samples $(\mathbf{x}, y, c)$ uniformly across all non-empty groups $g$.
\end{itemize}
Uniform Group Sampling is found to be critical for eX2L (See Appendix \ref{appendix:a}), as it prevents the model from learning the inherent label-confounder proportions as a shortcut, further forcing the algorithm to rely on the Grad-CAM penalty to distinguish between invariant and spurious features.

Furthermore, we distinguish these from \textbf{Uniform Environment Sampling}, a method typically used in domain generalization to sample uniformly from predefined environments. We do not employ environment-based sampling as $\hat{R}_\text{eX2L}$ is specifically designed for the group-based setting, focusing more on the label and confounder intersection of attributes. The complete training procedure is summarized in Algorithm~\ref{alg:ex2l}.

\begin{algorithm}[tb]
  \caption{eXplaining to Learn (eX2L)}
  \label{alg:ex2l}
  \begin{algorithmic}
    \STATE {\bfseries Input:} Training set $\mathcal{D}$, label model $m_\theta$, confounder model $m_\phi$, loss functions $\mathcal{L}$, similarity metric $S$, hyperparameters $\lambda_c, \lambda_{sim}$.
    \STATE {\bfseries Initialize:} Weights $\theta, \phi$ and optimizers.
    \FOR{each epoch}
        \FOR{batch $(\mathbf{x}, y, c) \in \mathcal{D}$}
            \STATE // \textit{Forward Pass}
            \STATE Pass $\mathbf{x}$ through models to get logit arrays:
            \STATE $\mathbf{s}_y \leftarrow m_\theta(\mathbf{x})$, $\mathbf{s}_c \leftarrow m_\phi(\mathbf{x})$

            \STATE // \textit{Targeted Logit Selection}
            \STATE Select actual target-class logits:
            \STATE $\hat{y}^{y} \leftarrow \mathbf{s}_y[y]$ if $|\mathcal{Y}|>2$, else $(2y-1)\cdot\mathbf{s}_y[0]$
            \STATE $\hat{c}^{c} \leftarrow \mathbf{s}_c[c]$ if $|C|>2$, else $(2c-1)\cdot\mathbf{s}_c[0]$
            \STATE // \textit{Grad-CAM Heatmap Generation}
            \STATE Extract feature maps $A_y, A_c$ from last conv layers.
            \FOR{$k \in \{1, \dots, |A_y|\}$}
                \STATE Extract and detach gradients per activation map.
                \STATE $\alpha^{y}_{k} \leftarrow \frac{1}{P} \sum_{i}\sum_{j} \frac{\partial{\hat{y}^y}}{\partial{A^{k}_{y,i, j}}}$
                \STATE $\alpha^{c}_{k} \leftarrow \frac{1}{P} \sum_{i}\sum_{j} \frac{\partial{\hat{c}^c}}{\partial{A^{k}_{c,i, j}}}$
            \ENDFOR
            \STATE $L^{y}_{\text{Grad-CAM}} \leftarrow \text{MinMax}_{\text{sample}}(\text{ReLU}(\sum_{k} \alpha^{y}_{k}A^{k}))$
            \STATE $L^{c}_{\text{Grad-CAM}} \leftarrow \text{MinMax}_{\text{sample}}(\text{ReLU}(\sum_{k} \alpha^{c}_{k}A^{k}))$
            \STATE // \textit{Objective Function}
            \STATE $\mathcal{L}_{task} \leftarrow \mathcal{L}(\mathbf{s}_y, y) + \lambda_c \mathcal{L}(\mathbf{s}_c, c)$
            \STATE // \textit{Similarity of Grad-CAM Heatmaps}
            \STATE $\mathcal{L}_{sim} \leftarrow S(L^y_{Grad-CAM}, L^c_{Grad-CAM})$
            \STATE $\mathcal{L}_{total} \leftarrow \mathcal{L}_{task} + \lambda_{sim} \mathcal{L}_{sim}$

            \STATE // \textit{Optimization}
            \STATE Update $\theta, \phi$ via $\nabla_{\theta, \phi} \mathcal{L}_{total}$
        \ENDFOR
    \ENDFOR
  \end{algorithmic}
\end{algorithm}

\subsection{Theoretical Motivation}

The design of eX2L is grounded in three key observations from the explainability and representation learning literature.

\paragraph{Grad-CAM as a Faithful Localization Mechanism.}
Crucially, \citet{adebayo2018sanity} demonstrated that unlike backpropagation-based attribution methods, which remain largely invariant to network reparameterization and behave similarly to edge detectors, gradient-based methods like Grad-CAM produce \textit{target-specific} attributions that faithfully reflect the model's learned parameters. This property makes Grad-CAM, in comparison to other attribution methods, well-suited for identifying which spatial regions drive predictions for both the label $y$ and the confounder $c$ throughout training.

\paragraph{Disjoint Attention Induces Feature Decorrelation.}
Let $\mathcal{A}_y \subseteq \mathcal{I}$ and $\mathcal{A}_c \subseteq \mathcal{I}$ denote the spatial regions attended to by the label and confounder models, respectively, where $\mathcal{I}$ represents the image domain. By minimizing the similarity $S(L^{y}_{\text{Grad-CAM}}, L^{c}_{\text{Grad-CAM}})$, we encourage $|\mathcal{A}_y \cap \mathcal{A}_c|$ to be small. This spatial disjointness discourages the label classifier from relying on confounding regions, effectively forcing it to discover alternative features that are predictive of $y$ but not of $c$.

This mechanism can be viewed as inducing a form of \textit{functional disentanglement}: rather than requiring full statistical independence of latent factors, which is provably difficult without inductive biases \citep{locatello2019challenging}, eX2L leverages explicit confounder supervision to enforce a weaker but practically sufficient objective where the label prediction $f_y$ becomes invariant to the spatial support of the confounder.

\paragraph{Connection to Information-Theoretic Principles.}
The information bottleneck principle \citep{tishby2015deep} posits that an optimal representation $Z$ should maximize information about the target $\mathcal{Y}$ while minimizing information about the input $\mathcal{X}$ that is irrelevant to $\mathcal{Y}$.

While we do not directly optimize an information-theoretic objective, the geometric constraint on Grad-CAM overlap serves as a computationally tractable proxy. By enforcing spatial disjointness between label and confounder attention maps, eX2L acts as a regularizer that reduces the model's capacity to exploit confounding features and encourages the model to rely on invariant features that generalize across groups.

\section{Experimental Setup}

\begin{table*}[t]
\centering
\small 
\setlength{\tabcolsep}{3.5pt} 

\caption{CMNIST and Waterbirds Experiment Results}
\label{table:1}
\begin{tabular}{@{}llcccc@{}}
\toprule
\multicolumn{2}{c}{\bf Algorithms} & \multicolumn{4}{c}{\bf Subpopulation Shift} \\
\cmidrule(lr){3-6}
\multirow{2}{*}{\bf Name} & \multirow{2}{*}{\bf Sampling} & \multicolumn{2}{c}{\bf CMNIST} & \multicolumn{2}{c}{\bf Waterbirds} \\
\cmidrule(lr){3-4} \cmidrule(lr){5-6}
& & \bf AA & \bf WGA & \bf AA & \bf WGA \\
\midrule
\multicolumn{6}{l}{\it Baselines} \\
ERM & Random & 32.96 $\pm$2.9\% & 25.11 $\pm$2.5\% & 62.86 $\pm$15.8\% & 23.68 $\pm$21.1\% \\
IRM & Uniform Env. & 65.11 $\pm$1.1\% & 64.59 $\pm$0.9\% & 89.16 $\pm$2.4\% & 84.42 $\pm$3.0\% \\
MMD & Uniform Env. & 65.41 $\pm$0.8\% & 61.98 $\pm$1.4\% & 90.81 $\pm$1.1\% & 85.64 $\pm$0.2\% \\
CORAL & Uniform Env. & 63.62 $\pm$4.1\% & 59.23 $\pm$7.0\% & 90.00 $\pm$0.8\% & 85.41 $\pm$0.8\% \\
DANN & Uniform Env. & 67.32 $\pm$1.6\% & 66.42 $\pm$1.6\% & 90.71 $\pm$0.6\% & 85.41 $\pm$0.8\% \\
CDANN & Uniform Env. & 35.25 $\pm$24.4\% & 21.00 $\pm$18.8\% & 89.57 $\pm$1.7\% & 84.71 $\pm$1.8\% \\
GroupDRO & Uniform Group & 66.14 $\pm$2.1\% & 63.79 $\pm$3.5\% & 90.50 $\pm$0.4\% & 85.95 $\pm$1.2\% \\
\midrule
\multicolumn{6}{l}{\it Ours} \\
\bf{eX2L-MAE} & \bf{Uniform Group} & \bf{69.02 $\pm$0.5\%} & \bf{67.63 $\pm$1.4\%} & 89.59 $\pm$0.5\% & \bf{87.45 $\pm$0.6\%} \\
eX2L-Cosine & Uniform Group & 67.90 $\pm$0.6\% & 66.15 $\pm$1.0\% & 90.85 $\pm$1.0\% & 85.72 $\pm$0.7\% \\
\bf{eX2L-Soft Dice} & \bf{Uniform Group} & 66.75 $\pm$1.6\% & 65.14 $\pm$2.1\% & \bf{92.12 $\pm$0.7\%} & \bf{86.92 $\pm$0.6\%} \\
\bf{eX2L-JS Dist.} & \bf{Uniform Group} & \bf{69.66 $\pm$0.9\%} & \bf{66.88 $\pm$2.3\%} & 89.18 $\pm$1.9\% & 85.10 $\pm$2.3\% \\
eX2L-SSIM & Uniform Group & 67.85 $\pm$1.0\% & 65.87 $\pm$2.1\% & 89.77 $\pm$1.5\% & 85.10 $\pm$1.2\% \\
\bottomrule
\end{tabular}

\vspace{20pt} 

\caption{Spawrious Experiment Results}
\label{table:2}
\begin{tabular}{@{}llcccc@{}}
\toprule
\multicolumn{2}{c}{\bf Algorithms} & \multicolumn{4}{c}{\bf Domain Generalization} \\
\cmidrule(lr){3-6}
\multirow{2}{*}{\bf Name} & \multirow{2}{*}{\bf Sampling} & \multicolumn{2}{c}{\bf Spawrious O2O (Easy)} &  \multicolumn{2}{c}{\bf Spawrious M2M (Hard)} \\
\cmidrule(lr){3-4} \cmidrule(lr){5-6}
& & \bf AA & \bf WGA & \bf AA & \bf WGA \\
\midrule
\multicolumn{6}{l}{\it Baselines} \\
ERM & Random & 91.00 $\pm$1.1\% & 84.12 $\pm$2.0\% & 77.37 $\pm$2.4\% & 54.91 $\pm$2.5\% \\
IRM & Uniform Env. & 93.01 $\pm$1.3\% & 86.20 $\pm$3.1\% & 64.66 $\pm$0.5\% & 35.19 $\pm$2.3\% \\
MMD & Uniform Env. & 93.13 $\pm$1.5\% & 86.17 $\pm$2.9\% & 76.64 $\pm$3.9\% & 54.35 $\pm$4.7\% \\
CORAL & Uniform Env. & 93.14 $\pm$1.6\% & 86.43 $\pm$3.0\% & 76.74 $\pm$3.5\% & 53.74 $\pm$4.7\% \\
DANN & Uniform Env. & 94.04 $\pm$1.0\% & 87.55 $\pm$1.2\% & 76.75 $\pm$4.8\% & 55.41 $\pm$9.3\% \\
CDANN & Uniform Env. & 93.43 $\pm$1.7\% & 85.24 $\pm$2.9\% & 73.95 $\pm$9.3\% & 48.23 $\pm$12.4\% \\
\bf{GroupDRO} & \bf{Uniform Group} & \bf{95.13 $\pm$0.3\%} & \bf{90.32 $\pm$1.0\%} & 77.70 $\pm$2.2\% & 54.40 $\pm$1.4\% \\
\midrule
\multicolumn{6}{l}{\it Ours} \\
\bf{eX2L-MAE} & \bf{Uniform Group} & \bf{94.30 $\pm$1.0\%} & \bf{88.05 $\pm$1.1\%} & \bf{82.24 $\pm$3.9\%} & \bf{66.31 $\pm$8.7\%} \\
eX2L-Cosine & Uniform Group & 93.97 $\pm$0.3\% & 87.71 $\pm$1.0\% & 81.37 $\pm$7.5\% & 64.25 $\pm$13.3\% \\
eX2L-Soft Dice & Uniform Group & 91.34 $\pm$1.2\% & 81.78 $\pm$2.1\% & 76.10 $\pm$4.5\% & 54.09 $\pm$7.8\% \\
eX2L-JS Dist. & Uniform Group & 93.81 $\pm$1.2\% & 87.74 $\pm$2.5\% & 60.77 $\pm$24.1\% & 38.02 $\pm$36.1\% \\
eX2L-SSIM & Uniform Group & 93.52 $\pm$1.0\% & 85.96 $\pm$2.4\% & 74.16 $\pm$6.6\% & 46.10 $\pm$11.9\% \\
\bottomrule
\end{tabular}
\end{table*}

\subsection{Datasets and Benchmarks}
We evaluate eX2L across two primary challenges: subpopulation shifts and domain generalization.

\paragraph{Subpopulation Shifts} We utilize standard benchmarks including \textit{CMNIST} \citep{arjovsky2020invariantriskminimization}, \textit{Waterbirds}, and \textit{CelebA} \citep{Sagawa*2020Distributionally}. These datasets allow us to test eX2L on synthetic subject-inherent correlations (e.g., color in CMNIST), synthetic natural background correlations (e.g., land/water in Waterbirds), and real-world attribute correlations (e.g., gender in CelebA).

\paragraph{Domain Generalization} We employ the \textit{Spawrious} synthetic data benchmark \citep{lynch2025spawrious}, specifically the \textit{One-to-One Easy} and \textit{Many-to-Many Hard} challenges. These tasks evaluate eX2L's ability to generalize across shifting background confounders (e.g., $\text{Dirt} \rightarrow \text{Beach}$ vs. $\text{Dirt, Snow} \rightarrow \text{Beach, Jungle}$). For Spawrious, the confounder labels $C$ correspond to the background category provided by the dataset \citep{lynch2025spawrious}, and groups $g$ are defined as the Cartesian product of dog breed label and background confounder, enabling the group-based eX2L setup to apply directly.

\subsection{Model Architecture}
All experiments utilize a \textbf{ResNet-50} backbone initialized with \texttt{IMAGENET1K\_V2} weights. To generate Grad-CAM maps, we use the final convolutional block of the module, as it captures semantically rich features more effectively than initial layers \citep{Selvaraju_2019}. These heatmaps are then detached before backpropagation.

\subsection{Baselines}
We compare eX2L against Empirical Risk Minimization (ERM) as a general baseline. For subpopulation shifts, we use \textbf{GroupDRO} \citep{Sagawa*2020Distributionally}. For domain generalization, we compare against SOTA environment-based methods including \textbf{IRM}, \textbf{MMD}, \textbf{CORAL}, \textbf{DANN}, and \textbf{CDANN}. We prioritize regularization-based methods, over data augmentation and fine-tuning methods, to ensure an analogous comparison to the eX2L framework.

\subsection{eX2L Regularization and Strategy}
We initially screened 11 similarity functions under two sampling regimes (random and uniform group sampling), totaling 22 configurations. To prioritize robustness, the top three functions per benchmark (by Worst-Group Accuracy (WGA) on CMNIST and Waterbirds) were selected, yielding five unique functions for full evaluation: \textit{Negative Mean Absolute Error (MAE)}, \textit{Negative Jensen-Shannon distance}, \textit{Cosine Similarity}, \textit{Structural Similarity Index Measure (SSIM)}, and \textit{Soft Dice}.

\subsection{Implementation Details}
\paragraph{Data Generation} We extended the DomainBed \citep{gulrajani2021in} repository to explicitly retrieve confounder annotations for \textit{CMNIST} and \textit{Spawrious}. We modified the WILDS repository \citep{Koh2020WILDSAB} such as \textit{Waterbirds} and \textit{CelebA} to simulate subpopulation shifts by creating varied group proportions across environments. More details may be viewed at Appendix \ref{subsection:groupprop}.

\paragraph{Training Strategy} Models were trained using SGD with a batch size of 128. Hyperparameters were tuned via randomized search (20 trials) following the distributions in \citet{gulrajani2021in}. For eX2L-specific parameters ($\lambda_c$ and $\lambda_{sim}$), distribution bounds were determined via an initial grid search. We employed early stopping (patience = 10; \text{min\_delta} = $10^{-3}$) based on the mean of Average Accuracy (AA) and WGA.

\FloatBarrier

\subsection{Evaluation Criteria}
Following \citet{Koh2020WILDSAB}, we report the mean and standard deviation across three runs using different seeds (42, 8, 777). Following \citet{Sagawa*2020Distributionally}, we use \textbf{Average Accuracy (AA)} and \textbf{Worst-Group Accuracy (WGA)} to evaluate the performance trade-offs inherent in distribution shifts.

\subsection{Auxiliary Studies}
Beyond performance metrics, we conduct the following analyses:
\begin{itemize}
    \item \textbf{Visual Interpretability}: Visualizing feature attribution via Grad-CAM \citep{Selvaraju_2019}.
    \item \textbf{Latent Representation}: Qualitative latent space evaluation via UMAP \citep{mcinnes2020umapuniformmanifoldapproximation} and quantitative distribution shift measurement via linear MMD directly on latent representation output ($Z$) \citep{JMLR:v13:gretton12a}.
    \item \textbf{Efficiency}: Analyzing the training time vs. performance trade-off in complex settings using the \textit{Hard Spawrious} benchmark.
\end{itemize}

Refer to Appendix \ref{appendix:b} for additional details regarding the experimental setup.

\section{Results and Discussion}
\label{sec:results}

\begin{figure*}[t]
\centering
\setlength{\tabcolsep}{2.5pt}
\begin{tabular}{ccc}
\textbf{Baseline} & \textbf{Subpopulation Shift SOTA} & \textbf{Ours} \\
\textit{(ERM)} & \textit{(GroupDRO)} & \textit{(eX2L)}\\[6pt]
\bmvaHangBox{\includegraphics[width=0.28\linewidth]{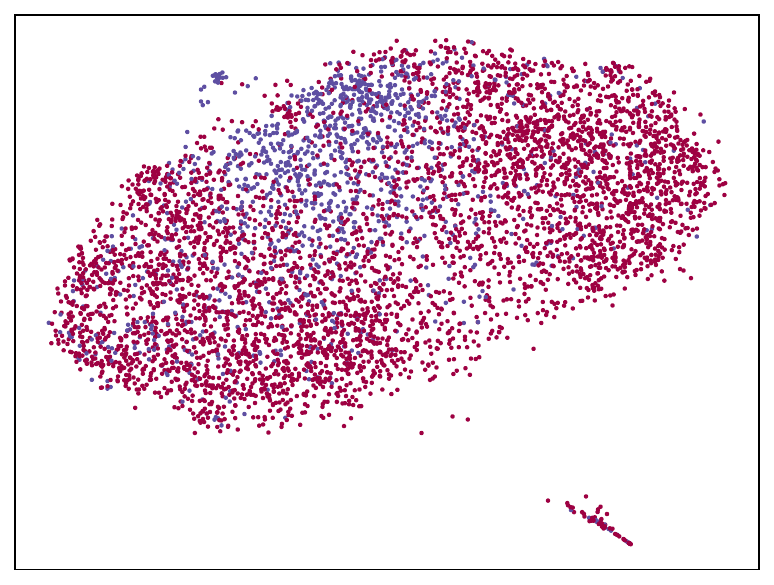}} &
\bmvaHangBox{\includegraphics[width=0.28\linewidth]{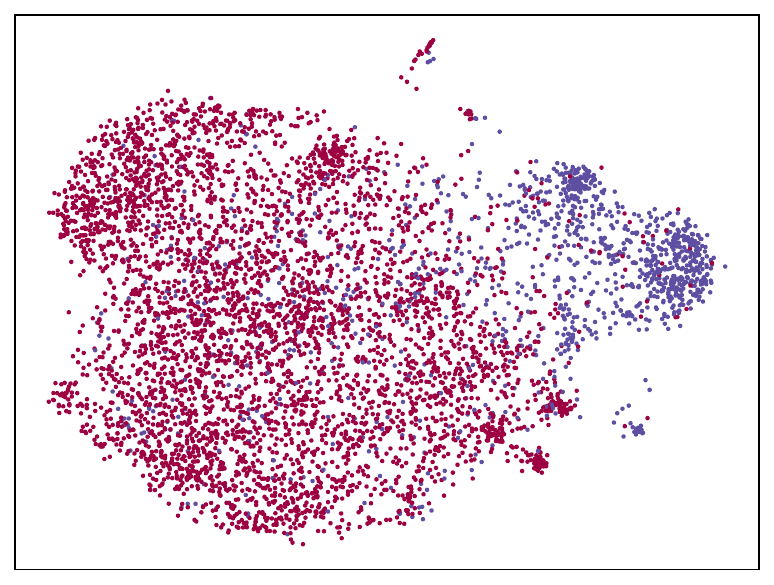}} &
\bmvaHangBox{\includegraphics[width=0.28\linewidth]{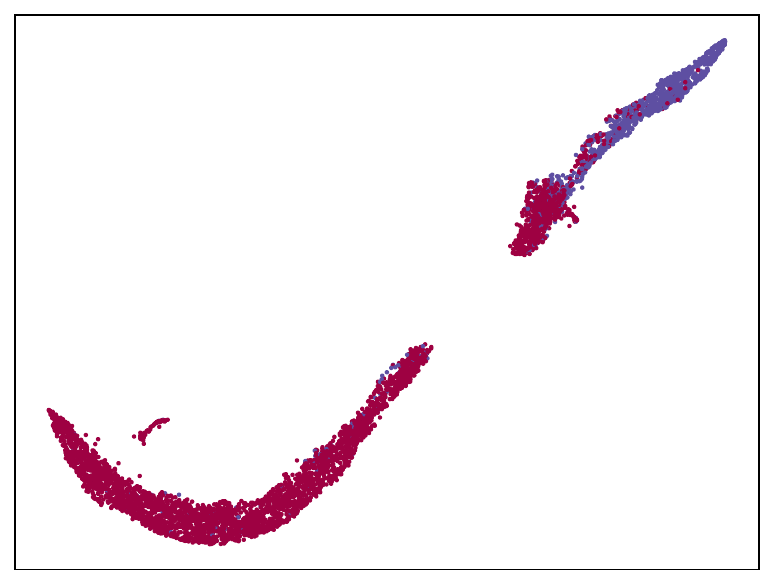}} \\
$\text{MMD}_{\sY} = 58.55$ & $\text{MMD}_{\sY} = 186.02$ & $\text{MMD}_{\sY} = 3598.97$  \\[6pt]
\multicolumn{3}{c}{(a) Label ($\sY$)}\\[6pt]
\bmvaHangBox{\includegraphics[width=0.28\linewidth]{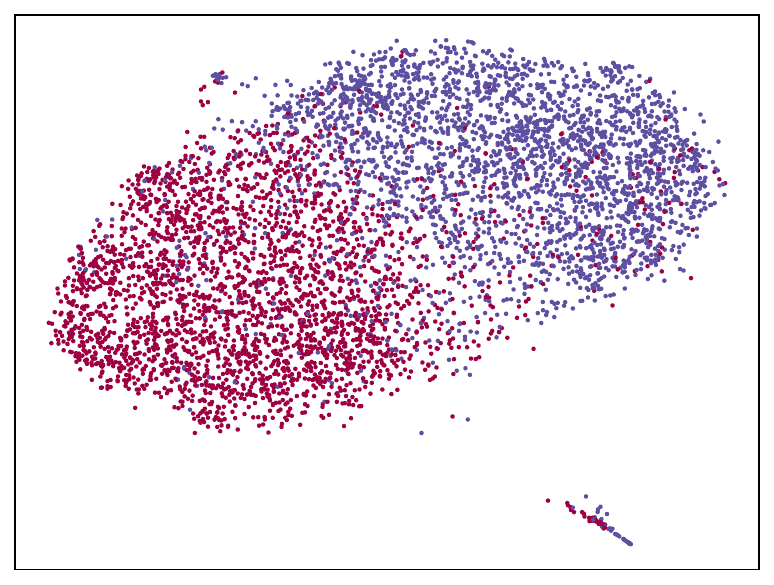}} &
\bmvaHangBox{\includegraphics[width=0.28\linewidth]{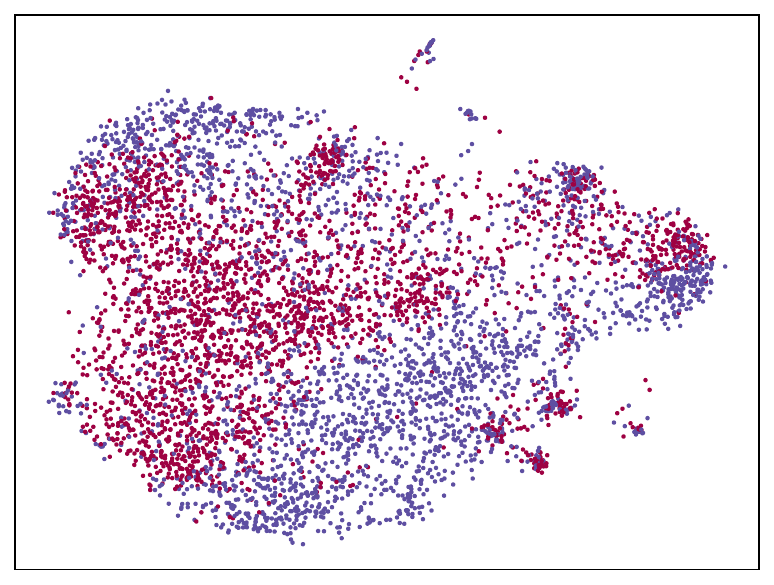}} &
\bmvaHangBox{\includegraphics[width=0.28\linewidth]{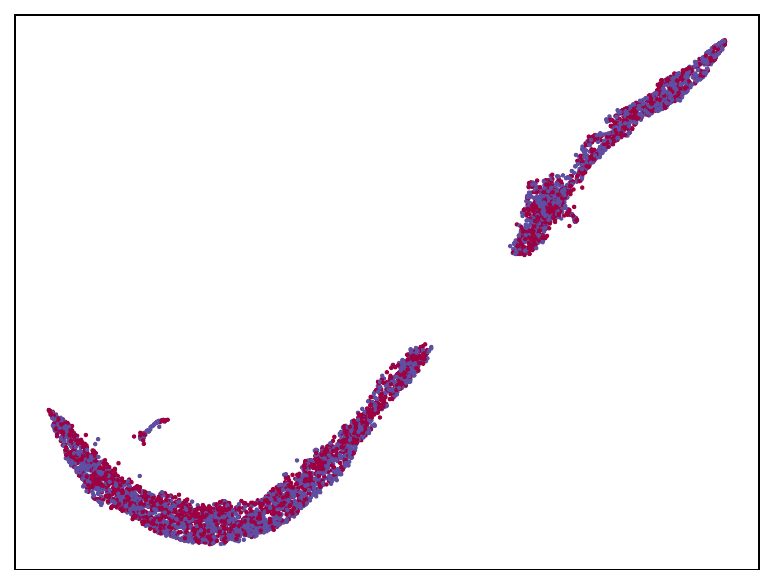}} \\
$\text{MMD}_{C} = 143.89$ & $\text{MMD}_{C} = 128.38$ & $\text{MMD}_{C} = 20.24$  \\[6pt]
\multicolumn{3}{c}{(b) Confounder ($C$)} \\[6pt]
\end{tabular}
\caption{Latent representation topology on the Waterbirds dataset mapped via UMAP. Points are color-coded by (a) true class label $\sY$ and (b) true background confounder $C$. ERM and GroupDRO display fractured, overlapping clusters heavily contaminated by background features, verified by high confounder divergence scores ($MMD_C = 143.89$ and $128.38$, respectively). Conversely, eX2L achieves functional feature disentanglement by collapsing confounder-predictive topology.}
\label{figure:2}
\end{figure*}

\begin{figure*}[t]
\centering
\setlength{\tabcolsep}{2.5pt}
\begin{tabular}{ccc}
\textbf{Baseline} & \textbf{Domain Generalization SOTA} & \textbf{Ours} \\
\textit{(ERM)} & \textit{(DANN)} & \textit{(eX2L)}\\[6pt]
\bmvaHangBox{\includegraphics[width=0.28\linewidth]{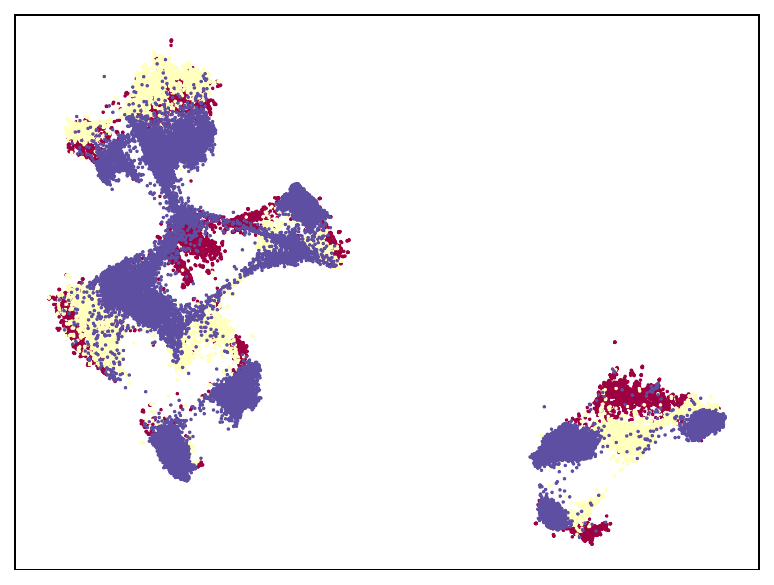}} &
\bmvaHangBox{\includegraphics[width=0.28\linewidth]{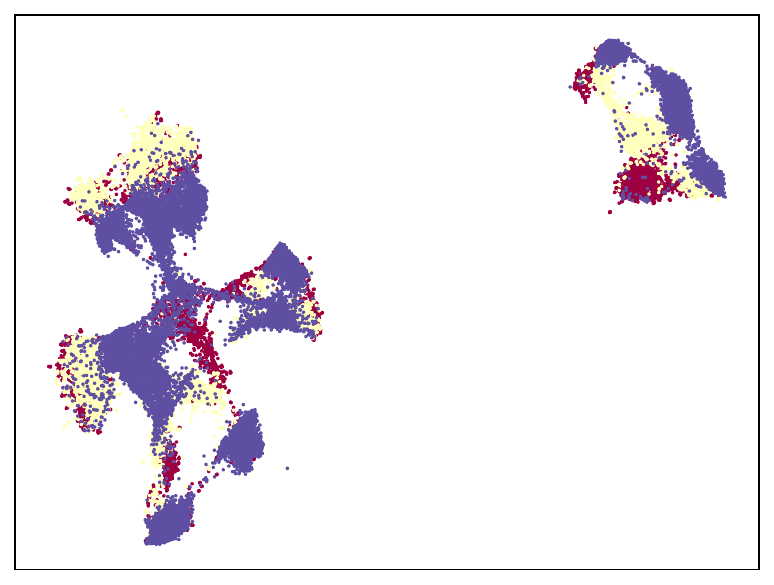}} &
\bmvaHangBox{\includegraphics[width=0.28\linewidth]{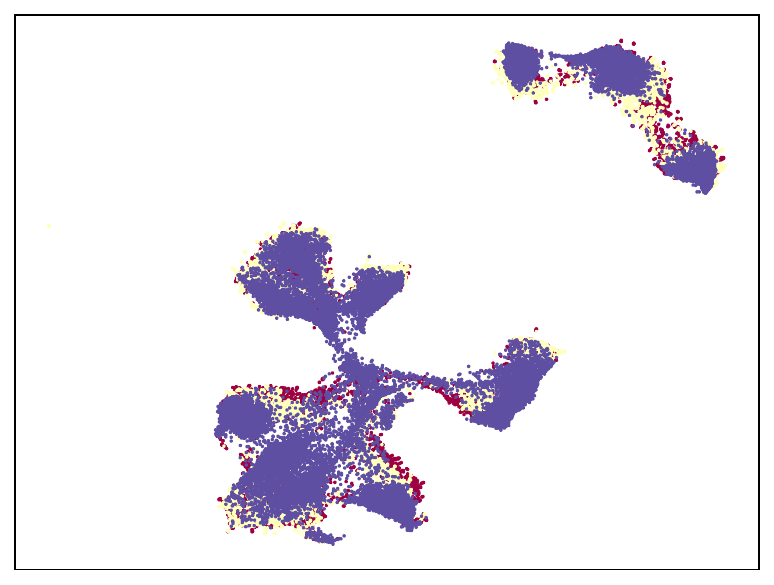}} \\
$\text{MMD}_{\text{Env}} = 19.12$ & $\text{MMD}_{\text{Env}} = 3.60$ & $\text{MMD}_{\text{Env}} = 2.79$  \\[6pt]
\end{tabular}
\caption{Emergent domain invariance under compositional shifts on the Spawrious Many-to-Many Hard dataset. Latent space coordinates are color-coded across three highly shifting training and deployment environments. Crucially, without requiring environmental annotations or domain adversarial training loops, eX2L exhibits superior distribution alignment. It yields the lowest global domain divergence ($MMD_{Env} = 2.79$, outperforming DANN's $3.60$ and ERM's $19.12$). This demonstrates that decoupling primary object attributes from local background nuisances at the group level serves as an elegant and highly effective computational proxy for traditional global domain generalization paradigms.}
\label{figure:3}
\end{figure*}

\subsection{Experiment Results}

\paragraph{Subpopulation Shift (Table~\ref{table:1})}
eX2L-based algorithms consistently outperform SOTA baselines in mitigating subpopulation shifts:
\begin{itemize}
    \item \textbf{eX2L-MAE Dominance:} eX2L-MAE achieves the highest WGA on \textit{CMNIST} ($67.63\%$) and competitive results on \textit{Waterbirds}, maintaining lower variance compared to baselines like \texttt{CDANN}.
    \item \textbf{Metric Sensitivity:} While eX2L-JS Dist. and eX2L-Soft Dice prioritize AA, eX2L-MAE is more effective for WGA. This suggests that MAE's stringent pixel-to-pixel penalty better prevents the model from latching onto spurious features.
\end{itemize}

\paragraph{Domain Generalization (Table~\ref{table:2})}
The \textit{Spawrious} results highlight a distinct bias-variance trade-off:
\begin{itemize}
    \item \textbf{Task Complexity:} While GroupDRO excels in the \textit{Easy (O2O)} setting, eX2L-MAE and eX2L-Cosine are the only algorithms to surpass $80\%$ AA and $60\%$ WGA in the \textit{Hard (M2M)} benchmark, led by eX2L-MAE.
    \item \textbf{Performance Gains:} In the \textit{Hard} setting, eX2L-MAE outperforms \texttt{DANN}, the strongest domain generalization baseline, by $5.49\%$ in AA and $10.90\%$ in WGA, and surpasses \texttt{GroupDRO}, the strongest overall baseline for AA ($77.70\%$), by $4.54\%$ in AA.
    \item \textbf{Robustness:} eX2L is uniquely robust to "many-to-many" correlations where multiple unseen backgrounds are present at test time, making it better suited for complex, real-world distribution shifts than existing regularization methods.
\end{itemize}

On \textit{CelebA}, eX2L-Cosine achieves $84.07\%$ WGA, matching GroupDRO's best WGA while maintaining $91.19\%$ AA; hair color's diffuse geometry favors softer metrics over pixel-wise penalties (see Appendix~\ref{appendix:c} and Appendix~\ref{appendix:f}).

\subsection{Visual Interpretability Analysis}

Based on Figure \ref{figure:1}, Grad-CAM heatmaps confirm that eX2L's gains stem from a conceptual shift in feature attribution:
\begin{itemize}
    \item \textbf{Structural Anchoring:} eX2L focuses on semantically invariant regions (e.g., digit edges for \textit{CMNIST}, ears for \textit{Dachshunds}), while baselines diffuse attention across spurious background cues.
\end{itemize}

\FloatBarrier

\subsection{Latent Representation Analysis}

\paragraph{Subpopulation Shift (Figure \ref{figure:2})}
The UMAP visualizations of the \textit{Waterbirds} dataset demonstrate that eX2L enforces superior feature disentanglement compared to SOTA baselines:
\begin{itemize}
    \item \textbf{Label Separation:} While ERM and GroupDRO exhibit overlapping or dispersed clusters, eX2L produces compact, well-separated representations with an exceptionally high $\text{MMD}_{\mathbb{Y}}$ ($3598.97$), indicating a highly precise decision boundary.
    \item \textbf{Confounder Independence:} Unlike ERM, which shows distinct confounder clusters (high $\text{MMD}_{C}$), eX2L achieves near-total interspersion of confounder features ($\text{MMD}_{C} = 20.24$). This suggests the model has successfully discarded spurious confounder signals in favor of class-relevant features.
\end{itemize}

\paragraph{Domain Generalization (Figure \ref{figure:3})}
The latent space on the \textit{Hard M2M Spawrious} dataset highlights the emergence of \textit{functional domain invariance}:
\begin{itemize}
    \item \textbf{Environmental Invariance:} eX2L exhibits the highest degree of environment interspersing, yielding the lowest $\text{MMD}_{\text{Env}}$ ($2.79$). Remarkably, this surpasses DANN ($3.60$), the domain generalization SOTA explicitly designed for this task.
    \item \textbf{Intrinsic Alignment:} eX2L aligns training and test distributions without requiring environmental labels or domain adversarial objectives.
\end{itemize}

\subsection{Efficiency Analysis}

\paragraph{Computational Efficiency vs. Performance (Table~\ref{table:3})}
The experimental results demonstrate that the added algorithmic complexity of eX2L is a necessary investment for superior generalization in difficult settings (full results in Appendix~\ref{appendix:d}):
\begin{itemize}
    \item \textbf{Complexity Pay-off:} While most baselines share ERM's time complexity, they often fail to exceed its performance. In contrast, eX2L-MAE and eX2L-Cosine achieve $1.21\times$ and $1.17\times$ better results, respectively, successfully converting additional training time into significant WGA gains.
    \item \textbf{Confounder Precision:} The \textit{Hard Spawrious} performance leap confirms that complex distribution shifts demand the precise confounder treatment eX2L provides.
\end{itemize}

\section{Conclusion}

We presented eX2L, an explanation-based framework that achieves functional domain invariance by enforcing spatial disjointness between label and confounder Grad-CAM maps. By providing a transparent mechanism to decouple spurious correlations, eX2L not only mitigates distribution shifts but challenges the perceived trade-off between interpretability and robustness, laying a foundation for algorithmic fairness.

\appendix

\section{Screening Process Results} \label{appendix:a}
\begin{table*}[t]
\caption{Similarity Function Selection Test Results}
\label{table:similarity}
\centering
\setlength{\tabcolsep}{4pt}
\begin{tabular}{@{}llcccc@{}}
\toprule
\multicolumn{2}{c}{\bf Algorithms} & \multicolumn{4}{c}{\bf Datasets} \\
\cmidrule(lr){3-6}
\multirow{2}{*}{\bf Name} & \multirow{2}{*}{\bf Sampling} & \multicolumn{2}{c}{\bf CMNIST} &  \multicolumn{2}{c}{\bf Waterbirds} \\
\cmidrule(lr){3-4} \cmidrule(lr){5-6}
& & \bf Train WGA & \bf Val WGA & \bf Train WGA & \bf Val WGA \\
\midrule
Cosine & Random & 42.51\% & 8.52\% & 15.04\% & 18.05\% \\
\bf{Cosine} & \bf{Uniform Group} & 72.38\% & 64.13\% & 91.55\% & \bf{85.71\%} \\
IoU & Random & 5.55\% & 0.84\% & 43.48\% & 44.64\% \\
IoU & Uniform Group & 65.67\% & 65.46\% & 91.70\% & 84.55\% \\
JS Dist. & Random & 0.00\% & 0.00\% & 25.10\% & 25.70\% \\
\bf{JS Dist.} & \bf{Uniform Group} & 71.31\% & \bf{67.83\%} & 88.20\% & 82.71\% \\
JS Div. & Random & 16.02\% & 8.75\% & 15.04\% & 19.55\% \\
JS Div. & Uniform Group & 71.19\% & 66.22\% & 90.68\% & 79.70\% \\
KL Div. & Random & 10.92\% & 5.29\% & 39.13\% & 45.11\% \\
KL Div. & Uniform Group & 68.44\% & 66.30\% & 89.41\% & 82.71\% \\
MAE & Random & 4.59\% & 5.38\% & 62.50\% & 53.38\% \\
\bf{MAE} & \bf{Uniform Group} & 72.20\% & 66.43\% & 89.93\% & \bf{85.71\%} \\
MSE & Random & 4.90\% & 2.09\% & 16.07\% & 18.05\% \\
MSE & Uniform Group & 64.07\% & 55.15\% & 87.08\% & 83.91\% \\
NCC & Random & 0.39\% & 0.30\% & 17.88\% & 18.05\% \\
NCC & Uniform Group & 72.37\% & 66.16\% & 91.60\% & 82.19\% \\
RMSE & Random & 17.74\% & 2.54\% & 10.71\% & 15.79\% \\
RMSE & Uniform Group & 65.52\% & 62.78\% & 91.09\% & 84.76\% \\
SSIM & Random & 74.77\% & 22.42\% & 14.87\% & 17.99\% \\
\bf{SSIM} & \bf{Uniform Group} & 71.48\% & \bf{67.41\%} & 89.97\% & 81.55\% \\
Soft Dice & Random & 15.34\% & 6.43\% & 46.43\% & 40.60\% \\
\bf{Soft Dice} & \bf{Uniform Group} & 69.79\% & \bf{66.52\%} & 93.06\% & \bf{84.96\%} \\
\bottomrule
\end{tabular}
\end{table*}

As shown in Table \ref{table:similarity}, algorithms which use Uniform Group Sampling performed better than their Random Sampling counterparts. This can be linked to how Uniform Group Sampling provides balanced views of the dataset per training batch, allowing the algorithm to focus more on decoupling the label and confounder attributes without the added noisiness of random batches.

\section{Additional Methodology Details} \label{appendix:b}
\subsection{Acquiring Target Logit in Binary Cases}
Given that binary models only provide single logits, to account for the two classes, the logit's additive inverse is used in the case that the target is negative or $0$. The following script is used where \texttt{targets} refers to the actual target variables:

\begin{lstlisting}
target_logit_for_grad = torch.where(targets.bool(), logits, -logits)
\end{lstlisting}

\subsection{Heatmap Detachment}
To avoid the computational costs associated with calculating gradient of gradients, during gradient calculation, \texttt{\text{create\_graph}} is set to \texttt{False} to detach the calculation of the heatmap from the backpropagation operation:

\begin{lstlisting}
gradients = torch.autograd.grad(outputs=target_logit_for_grad.sum(),
    inputs=activations['value'],
    retain_graph=True,
    create_graph=False)[0]
\end{lstlisting}

\subsection{Datasets}
\label{appendix:a.1}
\paragraph{CMNIST} CMNIST is a synthetic dataset based on the MNIST dataset, where the labels are divided into "Less than 5", and "Greater than or Equal to 5". \citep{arjovsky2020invariantriskminimization}. \citet{arjovsky2020invariantriskminimization} adds the spurious correlation, color, by coloring them red or green based on the label. This correlation is subverted at test time, where the spurious correlation is reversed \citep{arjovsky2020invariantriskminimization}. Additionally, labels are flipped with a 0.25 probability, to add noise in the dataset \citep{arjovsky2020invariantriskminimization}.

\paragraph{Waterbirds} Waterbirds is a synthetic dataset based on the CUB dataset, in which pictures of birds are spuriously correlated with the land or water background \citep{Sagawa*2020Distributionally}. The challenge here is that birds which do not necessarily match with their background (e.g., Water Bird in Land) are minority groups, leading to the problem of subpopulation shift at test time, where they are balanced out again \citep{Sagawa*2020Distributionally}.

\paragraph{CelebA} CelebA is based on the CelebA celebrity face dataset in which the labels are the hair color, and the confounders are the gender of the image subjects \citep{Sagawa*2020Distributionally}. The spurious correlation here is that most females are blondes, while most males are non-blondes \citep{Sagawa*2020Distributionally}.

\paragraph{Spawrious} Spawrious is a synthetic dataset created using generative AI images which tests models on unseen group combinations (dog breeds × backgrounds) at test time \citep{lynch2025spawrious}. It has two correlation types: 
\begin{itemize}
    \item \textbf{One-to-One}, where each breed is paired with one background during training (e.g., corgi → forest, dachshund → snow). A proportion from each dog breed is paired with the same background (e.g., n\% of corgi and dachshund → beach).
    \item \textbf{Many-to-Many}, where each breed is paired with two backgrounds (e.g., corgi → {snow, forest}).
\end{itemize}
    
In both correlation types, the pairings shift at test time. Each type has three difficulty levels (Easy, Medium, Hard) based on how well a standard ERM classifier performs on the respective train/test breed-background pairing combinations \citep{lynch2025spawrious}. 

To illustrate the difficulty levels, the backgrounds of \textbf{One-to-One}'s Easy Difficulty is comprised of desert, jungle, dirt, and snow, while its Hard difficulty has jungle, mountain, snow, and desert as its backgrounds. This is given by how a standard ERM classifier had an empirically lower performance in the background combinations of the Hard difficulty, as compared to the Easy difficulty.

All models were trained for 300 epochs for all datasets, with the exception of \textit{CelebA}, which was trained for 50 epochs due to its larger dataset size.

\subsection{Baseline Algorithms}
\subsubsection{Empirical Risk Minimization (ERM)}
Empirical Risk Minimization (ERM) only follows the standard cross-entropy loss such that in binary-class settings, the binary cross entropy (BCE) loss is used:

\begin{equation}
    \text{BCE} = -\frac{1}{n} \sum_{i = 1}^{n} \left(y_{i} \cdot \log \hat{y}_{i} + \left(1 - y_{i} \right) \cdot \log \left(1 - \hat{y}_{i} \right) \right)
\end{equation}

For multi-class settings with $K$ classes, the categorical cross-entropy (CCE) is used:
\begin{equation}
    \text{CCE} = -\frac{1}{n} \sum_{i = 1}^{n} \sum_{k=1}^{K} y_{i,k} \cdot \log \hat{y}_{i,k}
\end{equation}

where $y_{i,k} \in \{0,1\}$ is the indicator that sample $i$ belongs to class $k$.

\subsubsection{Invariant Risk Minimization (IRM) \texorpdfstring{\citep{arjovsky2020invariantriskminimization}}{}}
Invariant Risk Minimization (IRM) is a learning paradigm developed by \citet{arjovsky2020invariantriskminimization} which aims to find invariant, generalizable representations across multiple training environments.

Following \citet{arjovsky2020invariantriskminimization}, we used IRMv1, its more tractable form for the experiments. Instead of minimizing risk per environment, it minimizes the global environmental risk using a linear invariant predictor:

\begin{equation}
\mathcal{L}_{IRM, w=1.0} \left(\Phi^{\top} \right) = \sum_{e \in \mathcal{E}_\text {tr}} R^{e} \left(\Phi^{\top} \right) + \lambda \cdot \mathbb{D}_{\text{lin}} \left(1.0, \Phi^{\top}, e\right)
\end{equation}

where $w$ is a fixed scalar weight (set to $1.0$ in IRMv1), $\Phi^{\top}$ serves as the model's parameters, $R^{e}$ is the risk for a given environment $e$ in the set of all training environments $\mathcal{E}_{\text{tr}}$, and the IRMv1 penalty $\mathbb{D}_{\text{lin}}$ is the squared norm of the normal-equations residual, measuring how far the fixed classifier $w$ deviates from the environment-optimal linear predictor:

\begin{equation}
\begin{aligned}
\mathbb{D}_{\text{lin}}(w, \Phi, e) = \big\| &\mathbb{E}_{X^e} \left[ \Phi(X^e) \Phi(X^e)^\top \right] w \\
&- \mathbb{E}_{X^e, Y^e} \left[ \Phi(X^e) Y^e \right] \big\|^2
\end{aligned}
\end{equation}

A small value of $\mathbb{D}_{\text{lin}}(1.0, \Phi, e)$ indicates that $w{=}1.0$ is already near-optimal for environment $e$, which is the IRMv1 invariance condition.

\subsubsection{Maximum Mean Discrepancy (MMD) \texorpdfstring{\citep{li2018domain}}{}}
\citet{li2018domain}'s rationale for using MMD regularization is to learn feature representations which would generalize to unseen, non-training domains. To operate on the latent representation of each domain using a mean map operation $\mu\left(\cdot\right)$ to map these representations in the reproducing kernel Hilbert space (RKHS), $\mathcal{H}$, \citet{li2018domain} defines $\mu_P$:

\begin{equation}
\mu_P := \mu(P) = \mathbb{E}_{x \sim P}[\phi(x)] = \mathbb{E}_{x \sim P}[k(x, \cdot)]
\end{equation}

where $\phi : \mathbb{R}^d \to \mathcal{H}$ is a feature map, and $k(\cdot, \cdot)$ is the kernel function induced by $\phi(\cdot)$. Based on the MMD theory \citep{li2018domain}, the distance between the domains $l$ and $t$ (or $\mathbb{P}_{l}$ and $\mathbb{P}_{t}$), which serves as the MMD loss, can be measured by:

\begin{equation}
    \text{MMD}\left(\textbf{H}_{l}, \textbf{H}_{t}\right) = ||\mu_{\mathbb{P}_{l}} - \mu_{\mathbb{P}_{t}}||_{\mathcal{H}}
\end{equation}

\subsubsection{CORrelation ALignment (CORAL) \texorpdfstring{\citep{sun2016coral}}{}}
CORAL loss is simply the distance between the covariances $C$ of the source domain $S$ and target domain $T$ features \citep{sun2016coral}, as stated by:

\begin{equation}
\mathcal{L}_\text{CORAL} = \frac{1}{4d^{2}} ||C_{S} - C_{T}||^{2}_{F}
\end{equation}

where $d$ is the number of input dimensions and $||\cdot||^{2}_{F}$ is the squared matrix Frobenius norm. The covariance matrices of the source and target data are given by \citep{sun2016coral}:

\begin{equation}
C_S = \frac{1}{n_S - 1} \left( D_S^\top D_S - \frac{1}{n_S} (\mathbf{1}^\top D_S)^\top (\mathbf{1}^\top D_S) \right)
\end{equation}
\begin{equation}
C_T = \frac{1}{n_T - 1} \left( D_T^\top D_T - \frac{1}{n_T} (\mathbf{1}^\top D_T)^\top (\mathbf{1}^\top D_T) \right)
\end{equation}

where $\mathbf{1}$ is a column vector with all elements equal to 1.

\subsubsection{Domain Adversarial Neural Networks (DANN) \texorpdfstring{\citep{ganin2016domain}}{}}

In DANN, \citet{ganin2016domain} created an adversarial network consisting of a label classifier and a domain classifier, where the novel contribution of the authors is introducing the gradient reversal, in which the additive inverse of the gradient from the domain classifier is backpropagated to the common feature extractor. This leads to the domain classifier being confused by the features from the feature extractor, while enforcing domain invariance in the feature extractor.

This is encapsulated by the minimax objective which maximizes the loss of the domain classifier while minimizing the loss of the label classifier, to make the feature extractor's parameters $\theta_f$ domain invariant:

\begin{equation}
\begin{aligned}
E(\theta_f, \theta_y, \theta_d) ={} & \frac{1}{n} \sum_{i=1}^{n} \mathcal{L}^{i}_{y}(\theta_f, \theta_y) - \lambda\bigg(\frac{1}{n} \sum_{i=1}^{n} \mathcal{L}_{d}^{i} (\theta_f, \theta_d) \\
& \qquad\quad + \frac{1}{n'} \sum_{i=n+1}^{N} \mathcal{L}_{d}^{i} (\theta_f, \theta_d)\bigg)
\end{aligned}
\end{equation}

where $\theta_d$ and $\theta_y$ are the parameters of the domain and label classifiers, $N$ is the total number of data points, $n$ is the total number of data points in the source domain, $n'$ is the total number of data points in the target domain or $N - n$, and $\mathcal{L}_d$ and $\mathcal{L}_y$ are the cross-entropy losses of the label and domain classifiers, respectively.

\subsubsection{Conditional Domain Adversarial Neural Networks (CDANN) \texorpdfstring{\citep{li2018cdann}}{}}

\citet{li2018cdann} posited that adapting not only the feature extractor, but also the domain classifier on class label information can help in promoting domain invariance. By conditioning the domain classifier on the data point's class information, the joint distribution of features and class labels across domains can be aligned, without hampering performance by blindly aligning all feature distributions across domains as in DANNs \citep{li2018cdann}.

As compared to DANN whose domain classifier's function is defined as $d(f)$ where $f$ represents the features, CDANN defines its domain classifier function as $d(f, y)$ where $(f, y)$ serves as the joint variable for the features and the class information.

\subsubsection{Group Distributionally Robust Optimization (GroupDRO) \texorpdfstring{\citep{sagawa2020distributionallyrobustneuralnetworks}}{}}
In GroupDRO \citep{sagawa2020distributionallyrobustneuralnetworks}, the primary objective is to find the model which minimizes the risk of the worst group, using the intuition that addressing the generalization problem of the worst group would lead to generalizing to other groups.

\citet{sagawa2020distributionallyrobustneuralnetworks} defines the worst-group risk as:

\begin{equation}
\mathcal{R}(\theta) = \max_{g \in \mathcal{G}} \mathbb{E}_{(x, y) \sim P_{g}} [\mathcal{L}(\theta; (x, y))]
\end{equation}

Given this, the primary objective of GroupDRO is to find the model parameters $\hat{\theta}_{\text{DRO}}$ which minimizes the empirical worst-group risk $\hat{\mathcal{R}}(\theta)$:

\begin{equation}
\hat{\theta}_{\text{DRO}} := \argmin_{\theta \in \Theta} \left\{ \hat{\mathcal{R}}(\theta) := \max_{g \in \mathcal{G}} \mathbb{E}_{(x, y) \sim P_{g}} [\mathcal{L}(\theta; (x, y))] \right\}
\end{equation}

\subsection{Similarity Functions}
\subsubsection{Negative Mean Absolute Error (MAE)}

Given that MAE is a distance metric and is boundless, its additive inverse is used to penalize similarity:

\begin{equation}
\text{MAE}_{\text{eX2L}} = -\frac{1}{n} \sum_{i} |L_{i}^{y} - L_{i}^{c} |
\end{equation}

where $L^y$ and $L^c$ are Grad-CAM heatmaps of the label and confounder model and $n$ is the total number of heatmap ``pixels''.

\subsubsection{Negative Mean Squared Error (MSE)}

Similar to MAE, the additive inverse is also used for MSE:

\begin{equation}
\text{MSE}_{\text{eX2L}} = -\frac{1}{n} \sum_{i} (L_{i}^{y} - L_{i}^{c} )^{2}
\end{equation}

\subsubsection{Negative Root Mean Squared Error (RMSE)}

Similar to MAE and MSE, the additive inverse is also used for RMSE:

\begin{equation}
\text{RMSE}_{\text{eX2L}} = -\sqrt{\frac{1}{n} \sum_{i} (L_{i}^{y} - L_{i}^{c} )^{2}}
\end{equation}

\subsubsection{Soft Intersection-over-Union (IoU)}

Given that the original IoU metric uses non-differentiable operations given that it is defined as the intersection of two sets over their union, we used its approximation as defined by \citet{rahman2016iou}:

\begin{equation}
\text{Soft-IoU}_{\text{eX2L}} = \frac{\sum_{i=1}^{n}L_{i}^{y}\cdot L_{i}^{c}}{\sum_{i=1}^{n} (L_{i}^{y}+L_{i}^{c} -L_{i}^{y} \cdot L_{i}^{c})}
\end{equation}

By penalizing spatial intersections, we are then able to penalize confounder reliance in the label model.

\subsubsection{Negative Kullback-Leibler (KL) Divergence}

The KL Divergence is a non-symmetric metric which measures the difference in information between two distributions. In $D_{\text{KL}}(P \| Q)$, $P$ is the reference distribution and $Q$ is the approximating distribution; the divergence is large when $Q$ is a poor approximation of $P$. Here we set the confounder heatmap $P_{L^c_{L1}}$ as the reference and the label heatmap $P_{L^y_{L1}}$ as the approximating distribution, and use the additive inverse to maximize dissimilarity:

\begin{equation}
    D_{\text{KL}}(\text{P}_{L^{c}_{_{L1}}} || \text{P}_{L^{y}_{L1}})_{\text{eX2L}} = - \sum_{x \in \mathcal{X}} \text{P}_{L^{c}_{L1}}(x) \ln \frac{\text{P}_{L^{c}_{L1}}(x)}{\text{P}_{L^{y}_{L1}}(x)}
\end{equation}

where $\text{P}_{L^{c}_{L1}}$ refers to the $L_1$-normalized probability distribution of the confounder model's heatmap pixels, $\text{P}_{L^{y}_{L1}}$ refers to the $L_1$-normalized probability distribution of the label model's heatmap pixels, and $\mathcal{X}$ refers to the pixel index space.

$L1$ normalization is used to produce probability distributions out of the pixel maps and provide equal importance across all pixels within the pixel maps. Furthermore, the asymmetric choice is deliberate: a large $D_{\text{KL}}(P_{L^c_{L1}} \| P_{L^y_{L1}})$ means the label heatmap is a poor approximation of the confounder heatmap, confirming spatial disjointness. The converse direction, whether the confounder heatmap can approximate the label heatmap, is less informative for our objective because it may simply reflect insufficient training of the confounder model (e.g., under small $\lambda_c$).

\subsubsection{Negative Jensen-Shannon (JS) Divergence}

Given KL Divergence's limitation on having one distribution as a target and another as a source, we also used JS divergence, a symmetric form of the measuring distribution divergences. JS divergence is simply the average of KL divergence of two distributions to the mean of both distributions, where for eX2L, its additive inverse is used (to induce distributional dissimilarity):

\begin{equation}
\begin{aligned}
D_{\text{JS}}(\text{P}_{L^{c}_{L1}} || \text{P}_{L^{y}_{L1}})_{\text{eX2L}} ={} & -\frac{1}{2} D_\text{KL} \left(\text{P}_{L^{c}_{_{L1}}} \,\Big\Vert\, \tfrac{\text{P}_{L^{c}_{_{L1}}} + \text{P}_{L^{y}_{_{L1}}}}{2}\right) \\
& - \frac{1}{2} D_\text{KL}\left(\text{P}_{L^{y}_{_{L1}}} \,\Big\Vert\, \tfrac{\text{P}_{L^{c}_{_{L1}}} + \text{P}_{L^{y}_{_{L1}}}}{2}\right)
\end{aligned}
\end{equation}

Through this, we are able to simultaneously evaluate the reliance of both models on each other's supposed conceptual attributes.

\subsubsection{Negative Jensen-Shannon Distance (JSD)}

While JS Divergence shares some of the same qualities as JS Distance (JSD) such as being bounded and symmetric, JS Divergence does not satisfy the Triangle Inequality, leading to it not being a proper distance metric. Due to being the square root of the JS Divergence, it satisfies the Triangle Inequality, hence serving as a proper distance metric.

Similarly, its additive inverse is used to penalize similarities in distributions of the heatmaps:

\begin{equation}
\begin{aligned}
\text{JSD}_{\text{eX2L}} = -\bigg[ & \frac{1}{2} D_\text{KL} \left(\text{P}_{L^{c}_{_{L1}}} \,\Big\Vert\, \tfrac{\text{P}_{L^{c}_{_{L1}}}+ \text{P}_{L^{y}_{_{L1}}}}{2}\right) \\
& + \frac{1}{2} D_\text{KL}\left(\text{P}_{L^{y}_{_{L1}}} \,\Big\Vert\, \tfrac{\text{P}_{L^{c}_{_{L1}}} + \text{P}_{L^{y}_{_{L1}}}}{2}\right) \bigg]^{1/2}
\end{aligned}
\end{equation}

\subsubsection{Cosine Similarity}

In calculating the Cosine Similarity of the two heatmaps for eX2L, in comparison to the standard Cosine Similarity calculation, the heatmaps are first $L_1$-normalized to produce probability distributions from the heatmaps. These $L_1$-normalized heatmaps are then flattened to get the cosine similarity between the two heatmaps:

\begin{equation}
\cos \left(\theta\right)_{eX2L} = \frac{\sum_{i=1}^{n}L^{c}_{L1, i} \cdot L^{y}_{L1, i}} {\sqrt{\sum_{i=1}^{n} (L^{c}_{L1, i})^{2}} \cdot {\sqrt{\sum_{i=1}^{n} (L^{y}_{L1, i})^{2}}}}
\end{equation}

\subsubsection{Normalized Cross-Correlation (NCC) Coefficient}

Given that we do not need the lagging mechanism used in cross-correlation given that eX2L aims to compare the heatmaps coming from the same image, we used its zero-lag form, which can be simply defined as the Pearson Correlation Coefficient allowing us to compare each heatmap globally:

\begin{equation}
\text{NCC}_{\text{eX2L}} = \frac{\sum_{i=1}^{n}\left(L^{c}_{i} - \overline{L^{c}}\right) \cdot \left(L^{y}_{i} - \overline{L^{y}}\right)} {\sqrt{\sum_{i=1}^{n} \left(L^{c}_{i} - \overline{L^{c}}\right)^{2}} \cdot \sqrt{\sum_{i=1}^{n} \left(L^{y}_{i} - \overline{L^{y}}\right)^{2}}}
\end{equation}

where $\overline{L^{c}}$ and $\overline{L^{y}}$ are the mean of all heatmap pixel values for the confounder and label models, respectively.

\subsubsection{Structural Similarity Index Measure (SSIM) \texorpdfstring{\citep{wang2004ssim}}{}}

\citet{wang2004ssim}'s SSIM aims to quantify similarities in structural information using three indicators: luminance; contrast; and structure.

The luminance comparison function is given by the twice the product of the mean pixel intensities ($\overline{L^{c}}, \overline{L^{y}}$) of the two heatmaps divided by the sum of their squared mean pixel intensities \citep{wang2004ssim}. A stabilizing constant $C_1$ is added to both numerator and denominator to avoid division by zero:

\begin{equation}
l(L^{c}, L^{y}) = \frac{2\overline{L^{c}} \cdot\overline{L^{y}} + C_{1}} {(\overline{L^{c}} )^{2} + (\overline{L^{y}} )^{2} + C_{1}}
\end{equation}

where $C_{1}$ is defined as:

\begin{equation}
C_{1} = (K_{1} L)^{2}
\end{equation}

where $K_{1} \ll 1$ and $L$ is the dynamic range of the pixel values \citep{wang2004ssim}.

The contrast comparison function, on the other hand, is given by:

\begin{equation}
c(L^{c}, L^{y}) = \frac{2\sigma_{L^{y}} \sigma_{L^{c}} + C_{2}}{\sigma_{L^{y}}^{2} + \sigma_{L^{c}}^{2} + C_{2}}
\end{equation}

where $C_{2} = (K_{2} L)^{2}$ and $K_{2} \ll 1$.

Lastly, the structure comparison function is defined by:

\begin{equation}
s(L^{c}, L^{y}) = \frac{\sigma_{L^{y}L^{c}} + C_{3}}{\sigma_{L^{y}}  \sigma_{L^{c}} + C_{3}}
\end{equation}

where $C_{3} = (K_{3} L)^{2}$, $K_{3} \ll 1$. Following \citet{wang2004ssim}, in practice $C_3 = C_2/2$ (i.e., $K_3 = K_2/\sqrt{2}$). The sample covariance $\sigma_{L^{y}L^{c}}$ between $L^{y}$ and $L^{c}$ can be estimated using:

\begin{equation}
\sigma_{L^{y}L^{c}} = \frac{1}{n-1} \sum_{i=1}^{n} (L^{y}_{i} - \overline{L^{y}}) (L^{c}_{i} - \overline{L^{c}})
\end{equation}

Altogether, SSIM \citep{wang2004ssim} is defined as:

\begin{equation}
    \text{SSIM}(L^{c}, L^{y}) = [l(L^{c}, L^{y})]^{\alpha} \cdot [c(L^{c}, L^{y})]^{\beta} \cdot [s(L^{c}, L^{y})]^{\gamma}
\end{equation}

where $\alpha > 0$, $\beta > 0$, and $\gamma > 0 $ are parameters used to indicate the relative importance of each comparison function.

\subsubsection{Soft Dice}

Given that the original Dice coefficient which is twice the intersection of sets A and B divided by the sum of the cardinality of both sets is not differentiable by itself, we used a non-squared variant of \citet{milletari2016vnetfullyconvolutionalneural}'s objective based on the coefficient such that:

\begin{equation}
\text{Soft-Dice}_\text{eX2L} = \frac{2\sum_{i=1}^{n}L_{i}^{c} \cdot L_{i}^{y}}{\sum_{i=1}^{n} (L_{i}^{c}) + \sum_{i=1}^{n} (L_{i}^{y})}
\end{equation}

Ranging from 0 to 1, through eX2L, this similarity metric penalizes spatial intersections, similar to Soft IoU.

\subsection{Group Proportion Strategy}
\label{subsection:groupprop}
For datasets such as \textit{CMNIST} and \textit{Spawrious}, the existing strategy for environment splits designed by \citet{arjovsky2020invariantriskminimization} are used.

As per Table \ref{table:waterbirds_group_prop}, for \textit{Waterbirds}, we created environments that differ greatly in terms of group proportions, thereby simulating subpopulation shift and encouraging the learning of spurious correlations, without the aid of proper regularization.

\begin{table}[t]
\caption{\textit{Waterbirds} Group Proportions}
\label{table:waterbirds_group_prop}
\centering
\small
\setlength{\tabcolsep}{3pt}
\begin{tabular}{@{}ccccc@{}}
\toprule
\textbf{Env.} & \textbf{Confounder} & \textbf{Label} & \textbf{Prop.} & \textbf{Combined} \\
\midrule
Land-centric (1) & Land & Waterbird & 8.26\% & \multirow{2}{*}{70.06\%} \\
Land-centric (1) & Land & Landbird & 61.80\% & \\
Land-centric (1) & Water & Waterbird & 11.58\% & \multirow{2}{*}{29.95\%} \\
Land-centric (1) & Water & Landbird & 18.37\% & \\
\midrule
Balanced (2) & Land & Waterbird & 5.98\% & \multirow{2}{*}{50.75\%} \\
Balanced (2) & Land & Landbird & 44.77\% & \\
Balanced (2) & Water & Waterbird & 19.05\% & \multirow{2}{*}{49.25\%} \\
Balanced (2) & Water & Landbird & 30.20\% & \\
\bottomrule
\end{tabular}
\end{table}

Each environment is made more distinct in terms of group proportions to ensure that the domain generalization algorithms learn different shifts in subpopulations.

As per Table \ref{table:celeba_group_prop}, for \textit{CelebA}, we created group proportions that model differences in sex, to encourage the unregularized model to learn the spurious correlation of sex with the primary attribute, hair color.

\begin{table}[t]
\caption{\textit{CelebA} Group Proportions}
\label{table:celeba_group_prop}
\centering
\small
\setlength{\tabcolsep}{3pt}
\begin{tabular}{@{}ccccc@{}}
\toprule
\textbf{Env.} & \textbf{Confounder} & \textbf{Label} & \textbf{Prop.} & \textbf{Combined} \\
\midrule
Balanced (1) & Male & Blonde & 1.08\% & \multirow{2}{*}{48.02\%} \\
Balanced (1) & Male & Non-Blonde & 46.94\% & \\
Balanced (1) & Female & Blonde & 10.20\% & \multirow{2}{*}{51.97\%} \\
Balanced (1) & Female & Non-Blonde & 41.77\% & \\
\midrule
Less Males (2) & Male & Blonde & 0.66\% & \multirow{2}{*}{35.85\%} \\
Less Males (2) & Male & Non-Blonde & 35.19\% & \\
Less Males (2) & Female & Blonde & 17.36\% & \multirow{2}{*}{64.15\%} \\
Less Males (2) & Female & Non-Blonde & 46.79\% & \\
\bottomrule
\end{tabular}
\end{table}

Overall, two environments are created: an environment with more balanced group proportions and an environment with more imbalanced group proportions to ensure that the domain generalization algorithms learn different shifts in subpopulations.

\subsection{Initial Grid Search Parameters}

The values used for conducting the initial grid search are ${0.001, 0.01, 0, 1, 10, 100}$ for both $\lambda_{y}$ and $\lambda_{c}$. Based on the initial results from the grid search on \textit{CMNIST} and \textit{Waterbirds}, the performance is most optimal between 0.1 and 100, leading to the hyperparameter uniform distribution, $U(0.1, 100)$ for both $\lambda_y$ and $\lambda_c$.

\subsection{Hyperparameter Grid \texorpdfstring{\citep{gulrajani2021in}}{}}

To promote better comparability, the hyperparameters used for doing the randomized search are based on the following random distributions used by \citet{gulrajani2021in} in their DomainBed evaluation suite. Their table is appended to, to reflect faithfully both their hyperparameter registry and the hyperparameter distributions used for the eX2L experiments, as shown in Table \ref{tab:hyperparameters}.

\begin{table*}[t]
\centering
\caption{Hyperparameter Search Space and Default Values}
\label{tab:hyperparameters}
\small
\begin{tabular}{llll}
\toprule
\textbf{Condition} & \textbf{Parameter} & \textbf{Default Value} & \textbf{Random Distribution} \\ \midrule
\multirow{1}{*}{ResNet} & Batch size & $128$ & --- \\
\midrule
\multirow{7}{*}{MNIST} & Learning rate & $0.00005$ & $10^{\text{Uniform}(-5, -3.5)}$ \\
 & Label model learning rate & $0.00005$ & $10^{\text{Uniform}(-5, -3.5)}$ \\
 & Confounder model learning rate & $0.00005$ & $10^{\text{Uniform}(-5, -3.5)}$ \\
 & Gen. learning rate & $0.00005$ & $10^{\text{Uniform}(-5, -3.5)}$ \\
 & Disc. learning rate & $0.00005$ & $10^{\text{Uniform}(-5, -3.5)}$ \\
 & Weight decay & $0$ & $0$ \\
 & Gen. weight decay & $0$ & $0$ \\ \midrule
\multirow{7}{*}{Not MNIST}  & Learning rate & $0.001$ & $10^{\text{Uniform}(-4.5, -3.5)}$ \\
 & Label model learning rate & $0.001$ & $10^{\text{Uniform}(-4.5, -2.5)}$ \\
 & Confounder model learning rate & $0.001$ & $10^{\text{Uniform}(-4.5, -2.5)}$ \\
 & Gen. learning rate & $0.001$ & $10^{\text{Uniform}(-4.5, -2.5)}$ \\
 & Disc. learning rate & $0.001$ & $10^{\text{Uniform}(-4.5, -2.5)}$ \\
 & Weight decay & $0$ & $10^{\text{Uniform}(-6, -2)}$ \\
 & Gen. weight decay & $0$ & $10^{\text{Uniform}(-6, -2)}$ \\ \midrule
 \multirow{4}{*}{eX2L} & Lambda ($y$) & $1.0$ & $10^{\text{Uniform}(-1, 2)}$ \\
 & Lambda ($c$) & $1.0$ & $10^{\text{Uniform}(-1, 2)}$ \\
 & Label model weight decay & $0$ & $10^{\text{Uniform}(-6, -2)}$ \\
 & Confounder model weight decay & $0$ & $10^{\text{Uniform}(-6, -2)}$ \\ \midrule
\multirow{5}{*}{\shortstack[l]{DANN, \\ C-DANN}} & Lambda & $1.0$ & $10^{\text{Uniform}(-2, 2)}$ \\
 & Disc. weight decay & $0$ & $10^{\text{Uniform}(-6, -2)}$ \\
 & Disc. steps & $1$ & $2^{\text{Uniform}(0, 3)}$ \\
 & Gradient penalty & $0$ & $10^{\text{Uniform}(-2, 1)}$ \\ \midrule
IRM & Lambda & $100$ & $10^{\text{Uniform}(-1, 5)}$ \\
 & Penalty annealing iters. & $500$ & $10^{\text{Uniform}(0, 4)}$ \\ \midrule
GroupDRO & Eta & $0.01$ & $10^{\text{Uniform}(-1, 1)}$ \\ \midrule
MMD & Gamma & $1$ & $10^{\text{Uniform}(-1, 1)}$ \\ \bottomrule
\end{tabular}
\end{table*}

\subsection{Reproducibility}
Experiments were conducted using PyTorch 2.10 and CUDA 12.1 on NVIDIA A100 and L4 GPUs. We enforced deterministic CUDNN and Torch settings to ensure comparative validity.

The following script is used to ensure reproducibility, which also considers the usage of workers:
\begin{lstlisting}
os.environ["CUBLAS_WORKSPACE_CONFIG"] = ":4096:8"

def set_seed(seed):
    random.seed(seed)
    np.random.seed(seed)
    torch.manual_seed(seed)
    torch.cuda.manual_seed(seed)
    torch.cuda.manual_seed_all(seed)
    torch.backends.cudnn.deterministic = True
    torch.backends.cudnn.benchmark = False
    torch.use_deterministic_algorithms(True)
    os.environ['PYTHONHASHSEED'] = str(seed)

def seed_worker(worker_id):
    worker_seed = torch.initial_seed() % 2**32
    np.random.seed(worker_seed)
    random.seed(worker_seed)

set_seed(seed)
g = torch.Generator()
g.manual_seed(seed)

dataloader = DataLoader(dataset, batch_size=batch_size, shuffle=True,
    pin_memory=True, num_workers=num_workers,
    persistent_workers=True, worker_init_fn=seed_worker,
    prefetch_factor=2, generator=g,
)
\end{lstlisting}

\subsection{Choice of Explanation Method}
eX2L uses Grad-CAM \citep{Selvaraju_2019} as the explanation signal during training rather than input-gradient methods such as Integrated Gradients or SHAP. The key distinction, established by \citet{adebayo2018sanity}, is that many input-gradient methods produce explanations that remain invariant under network reparametrization: they function effectively as unsupervised edge detectors whose output reflects image structure more than learned model behavior. Grad-CAM, by contrast, is sensitive to the current parameter state of the network, making its output a faithful reflection of \textit{what the model has specifically learned to attend to} at each training step. This parameter-sensitivity is essential for eX2L: the $\mathcal{L}_{\text{sim}}$ penalty is designed to penalize overlap between model-specific visual attention patterns, not between shared structural features of the input image. Using an input-agnostic method would therefore conflate learned attention with image edges, undermining the debiasing signal. Extension to other parameter-sensitive attribution methods (e.g., GradCAM++ or Score-CAM) is a natural future direction and would not require architectural changes to eX2L.

\section{CelebA Results} \label{appendix:c}

On \textit{CelebA}, eX2L performs comparably to GroupDRO, the leading algorithm on this benchmark. The results reveal an important insight about the relationship between confounder geometry and similarity metric selection.

\paragraph{Why hair color is a harder confounder.} Background confounders (e.g., land/water in Waterbirds, breed-background pairs in Spawrious) produce spatially tight, high-contrast heatmap regions that are easily distinguished from label-relevant features. Hair color, by contrast, occupies diffuse, low-contrast areas of the face image that often overlap with structurally important regions (e.g., the top of the head, which is also part of the subject's identity). This means the label model and confounder model will inevitably share some activated pixels in the CelebA setting , not because of spurious learning, but because both the label (hair color) and the confounder (gender) involve the same physical region.

\paragraph{Effect on similarity metrics.} Strict pixel-wise penalties such as MAE apply uniform pressure across all pixels, over-penalizing semantically valid co-activations where the label feature and confounder feature genuinely co-occur in space. Softer similarity metrics (Cosine, Soft Dice) are more tolerant of partial overlap and better suited to this geometry. This explains the pattern in Table~\ref{table:celeba_exp_results}: eX2L-Cosine and eX2L-Soft Dice match GroupDRO's WGA ($84.07\%$) while maintaining competitive AA ($91.19\%$ and $91.70\%$), whereas eX2L-MAE underperforms in this setting.

\paragraph{Implications.} CelebA does not indicate a failure of the eX2L framework; it indicates that metric selection should be informed by confounder geometry. The ablation in Appendix~\ref{appendix:a} confirms this: metrics that perform best on background confounders (MAE) are not universally optimal, and practitioners should use the screening protocol to select the appropriate similarity function for their confounder type.

\begin{table}[ht]
\caption{CelebA Experiment Results}
\label{table:celeba_exp_results}
\centering
\small
\setlength{\tabcolsep}{3pt}
\begin{tabular}{@{}llcc@{}}
\toprule
\multicolumn{2}{c}{\bf Algorithms} & \multicolumn{2}{c}{\bf Subpopulation Shift} \\
\cmidrule(lr){3-4}
\multirow{2}{*}{\bf Name} & \multirow{2}{*}{\bf Sampling} & \multicolumn{2}{c}{\bf CelebA} \\
\cmidrule(lr){3-4}
& & \bf AA & \bf WGA \\
\midrule
\multicolumn{4}{l}{\it Baselines} \\
ERM & Random & 95.56 $\pm$0.0\% & 39.63 $\pm$2.6\% \\
IRM & Uniform Env. & 91.17 $\pm$0.0\% & 82.04 $\pm$1.8\% \\
MMD & Uniform Env. & 92.19 $\pm$0.3\% & 81.30 $\pm$1.4\% \\
CORAL & Uniform Env. & 92.20 $\pm$0.4\% & 82.41 $\pm$0.8\% \\
DANN & Uniform Env. & 92.23 $\pm$0.3\% & 80.93 $\pm$2.2\% \\
CDANN & Uniform Env. & 90.16 $\pm$0.2\% & 80.56 $\pm$2.2\% \\
\bf{GroupDRO} & \bf{Uniform Group} & \bf{91.94 $\pm$0.3\%} & \bf{84.07 $\pm$0.6\%} \\
\midrule
\multicolumn{4}{l}{\it Ours} \\
eX2L-MAE & Uniform Group & 90.79 $\pm$0.2\% & 82.59 $\pm$2.0\% \\
\bf{eX2L-Cosine} & \bf{Uniform Group} & \bf{91.19 $\pm$0.3\%} & \bf{84.07 $\pm$1.6\%} \\
\bf{eX2L-Soft Dice} & \bf{Uniform Group} & \bf{91.70 $\pm$0.3\%} & \bf{83.33 $\pm$1.0\%} \\
eX2L-JS Dist. & Uniform Group & 90.05 $\pm$0.5\% & 84.07 $\pm$1.7\% \\
eX2L-SSIM & Uniform Group & 91.39 $\pm$0.2\% & 80.93 $\pm$2.7\% \\
\bottomrule
\end{tabular}
\end{table}

\section{Other Efficiency Analysis Results} \label{appendix:d}

As seen in Tables \ref{table:3}, \ref{table:cmnist}, \ref{table:waterbirds}, \ref{table:celeba}, and \ref{table:spawrious_o2o_easy}, eX2L exhibits a time-performance trade-off, where the added time complexity provides a corresponding increase in performance relative to the baseline ERM.

\begin{table}[ht]
\caption{Time-Performance Trade-off in \textit{Hard Many-to-Many Spawrious} Benchmark}
\label{table:3}
\centering
\small
\setlength{\tabcolsep}{2.8pt}
\begin{tabular}{@{}lcccc@{}}
\toprule
\textbf{Algorithms} & $\textbf{time}_{\textbf{epoch}}\textbf{(s)}$ & $\textbf{WGA}$ & $\frac{\textbf{time}_{\textbf{epoch}}}{\textbf{time}_\textbf{epoch}^{\textbf{ERM}}}$ & $\frac{\textbf{WGA}}{\textbf{WGA}_\textbf{ERM}}$ \\
\midrule
\multicolumn{5}{l}{\it Baselines} \\
ERM & 207.45 & 54.91 & --- & --- \\
IRM & 207.28 & 35.19 & 1.00$\times$ & 0.64$\times$ \\
MMD & 207.71 & 54.35 & 1.00$\times$ & 0.99$\times$ \\
CORAL & 208.61 & 53.74 & 1.01$\times$ & 0.98$\times$ \\
DANN & 204.95 & 55.41 & 0.99$\times$ & 1.01$\times$ \\
CDANN & 209.49 & 48.23 & 1.01$\times$ & 0.88$\times$ \\
GroupDRO & 213.01 & 54.40 & 1.03$\times$ & 0.99$\times$ \\
\midrule
\multicolumn{5}{l}{\it Ours} \\
\textbf{eX2L-MAE} & 403.50 & \textbf{66.31} & 1.95$\times$ & \textbf{1.21\bm{$\times$}} \\
\textbf{eX2L-Cosine} & 403.76 & \textbf{64.25} & 1.95$\times$ & \textbf{1.17\bm{$\times$}} \\
eX2L-Soft Dice & 405.98 & 54.09 & 1.96$\times$ & 0.99$\times$ \\
eX2L-JS Dist. & 415.73 & 38.02 & 2.00$\times$ & 0.69$\times$ \\
eX2L-SSIM & 406.36 & 46.10 & 1.96$\times$ & 0.84$\times$ \\
\bottomrule
\end{tabular}
\end{table}

\begin{table}[ht]
\caption{Time-Performance Trade-off in \textit{CMNIST} Benchmark}
\label{table:cmnist}
\centering
\small 
\setlength{\tabcolsep}{2.8pt} 
\begin{tabular}{@{}lcccc@{}}
\toprule
\textbf{Algorithms} & $\textbf{time}_{\textbf{epoch}}\textbf{(s)}$ & $\textbf{WGA}$ & $\frac{\textbf{time}_{\textbf{epoch}}}{\textbf{time}_\textbf{epoch}^{\textbf{ERM}}}$ & $\frac{\textbf{WGA}}{\textbf{WGA}_\textbf{ERM}}$ \\
\midrule
\multicolumn{5}{l}{\it Baselines} \\
ERM & 47.42 & 25.11 & --- & --- \\
IRM & 47.43 & 64.59 & 1.00$\times$ & 2.57$\times$ \\
MMD & 47.56 & 61.98 & 1.00$\times$ & 2.47$\times$ \\
CORAL & 47.49 & 59.23 & 1.00$\times$ & 2.36$\times$ \\
DANN & 46.19 & 66.42 & 0.97$\times$ & 2.65$\times$ \\
CDANN & 47.05 & 21.00 & 0.99$\times$ & 0.84$\times$ \\
GroupDRO & 49.66 & 63.79 & 1.05$\times$ & 2.54$\times$ \\
\midrule
\multicolumn{5}{l}{\it Ours} \\
\textbf{eX2L-MAE} & 93.28 & \textbf{67.63} & 1.97$\times$ & \textbf{2.69$\times$} \\
eX2L-JS Dist. & 93.36 & 66.88 & 1.97$\times$ & 2.66$\times$ \\
eX2L-Cosine & 93.96 & 66.15 & 1.98$\times$ & 2.63$\times$ \\
eX2L-SSIM & 95.67 & 65.87 & 2.02$\times$ & 2.62$\times$ \\
eX2L-Soft Dice & 93.33 & 65.14 & 1.97$\times$ & 2.59$\times$ \\
\bottomrule
\end{tabular}
\end{table}

\begin{table}[hb]
\caption{Time-Performance Trade-off in \textit{Waterbirds} Benchmark}
\label{table:waterbirds}
\centering
\small 
\setlength{\tabcolsep}{2.8pt} 
\begin{tabular}{@{}lcccc@{}}
\toprule
\textbf{Algorithms} & $\textbf{time}_{\textbf{epoch}}\textbf{(s)}$ & $\textbf{WGA}$ & $\frac{\textbf{time}_{\textbf{epoch}}}{\textbf{time}_\textbf{epoch}^{\textbf{ERM}}}$ & $\frac{\textbf{WGA}}{\textbf{WGA}_\textbf{ERM}}$ \\
\midrule
\multicolumn{5}{l}{\it Baselines} \\
ERM & 14.05 & 23.68 & --- & --- \\
IRM & 13.58 & 84.42 & 0.97$\times$ & 3.57$\times$ \\
MMD & 13.91 & 85.64 & 0.99$\times$ & 3.62$\times$ \\
CORAL & 13.69 & 85.41 & 0.97$\times$ & 3.61$\times$ \\
DANN & 13.69 & 85.41 & 0.97$\times$ & 3.61$\times$ \\
CDANN & 13.61 & 84.71 & 0.97$\times$ & 3.58$\times$ \\
GroupDRO & 14.34 & 85.95 & 1.02$\times$ & 3.63$\times$ \\
\midrule
\multicolumn{5}{l}{\it Ours} \\
\textbf{eX2L-MAE} & 28.45 & \textbf{87.45} & 2.02$\times$ & \textbf{3.69$\times$} \\
eX2L-Soft Dice & 28.54 & 86.92 & 2.03$\times$ & 3.67$\times$ \\
eX2L-Cosine & 28.68 & 85.72 & 2.04$\times$ & 3.62$\times$ \\
eX2L-JS Dist. & 27.70 & 85.10 & 1.97$\times$ & 3.59$\times$ \\
eX2L-SSIM & 28.79 & 85.10 & 2.05$\times$ & 3.59$\times$ \\
\bottomrule
\end{tabular}
\end{table}

\begin{table}[ht]
\caption{Time-Performance Trade-off in \textit{CelebA} Benchmark}
\label{table:celeba}
\centering
\small 
\setlength{\tabcolsep}{2.8pt} 
\begin{tabular}{@{}lcccc@{}}
\toprule
\textbf{Algorithms} & $\textbf{time}_{\textbf{epoch}}\textbf{(s)}$ & $\textbf{WGA}$ & $\frac{\textbf{time}_{\textbf{epoch}}}{\textbf{time}_\textbf{epoch}^{\textbf{ERM}}}$ & $\frac{\textbf{WGA}}{\textbf{WGA}_\textbf{ERM}}$ \\
\midrule
\multicolumn{5}{l}{\it Baselines} \\
ERM & 207.39 & 39.63 & --- & --- \\
IRM & 208.03 & 82.04 & 1.00$\times$ & 2.07$\times$ \\
MMD & 208.47 & 81.30 & 1.01$\times$ & 2.05$\times$ \\
CORAL & 208.53 & 82.41 & 1.01$\times$ & 2.08$\times$ \\
DANN & 205.12 & 80.93 & 0.99$\times$ & 2.04$\times$ \\
CDANN & 207.66 & 80.56 & 1.00$\times$ & 2.03$\times$ \\
GroupDRO & 213.10 & 84.07 & 1.03$\times$ & 2.12$\times$ \\
\midrule
\multicolumn{5}{l}{\it Ours} \\
\textbf{eX2L-Cosine} & 403.76 & \textbf{84.07} & 1.95$\times$ & \textbf{2.12$\times$} \\
\textbf{eX2L-JS Dist.} & 406.36 & \textbf{84.07} & 1.96$\times$ & \textbf{2.12$\times$} \\
eX2L-Soft Dice & 405.98 & 83.33 & 1.96$\times$ & 2.10$\times$ \\
eX2L-MAE & 403.50 & 82.59 & 1.95$\times$ & 2.08$\times$ \\
eX2L-SSIM & 415.73 & 80.93 & 2.00$\times$ & 2.04$\times$ \\
\bottomrule
\end{tabular}
\end{table}

\begin{table}[ht]
\caption{Time-Performance Trade-off in \textit{Easy One-to-One Spawrious} Benchmark}
\label{table:spawrious_o2o_easy}
\centering
\small 
\setlength{\tabcolsep}{2.8pt} 
\begin{tabular}{@{}lcccc@{}}
\toprule
\textbf{Algorithms} & $\textbf{time}_{\textbf{epoch}}\textbf{(s)}$ & $\textbf{WGA}$ & $\frac{\textbf{time}_{\textbf{epoch}}}{\textbf{time}_\textbf{epoch}^{\textbf{ERM}}}$ & $\frac{\textbf{WGA}}{\textbf{WGA}_\textbf{ERM}}$ \\
\midrule
\multicolumn{5}{l}{\it Baselines} \\
ERM & 26.58 & 84.12 & --- & --- \\
IRM & 26.87 & 86.20 & 1.01$\times$ & 1.02$\times$ \\
MMD & 26.97 & 86.17 & 1.01$\times$ & 1.02$\times$ \\
CORAL & 27.00 & 86.43 & 1.02$\times$ & 1.03$\times$ \\
DANN & 26.49 & 87.55 & 1.00$\times$ & 1.04$\times$ \\
CDANN & 26.61 & 85.24 & 1.00$\times$ & 1.01$\times$ \\
GroupDRO & 28.23 & 90.32 & 1.06$\times$ & 1.07$\times$ \\
\midrule
\multicolumn{5}{l}{\it Ours} \\
\textbf{eX2L-MAE} & 52.16 & \textbf{88.05} & 1.96$\times$ & \textbf{1.05$\times$} \\
eX2L-JS Dist. & 52.16 & 87.74 & 1.96$\times$ & 1.04$\times$ \\
eX2L-Cosine & 52.12 & 87.71 & 1.96$\times$ & 1.04$\times$ \\
eX2L-SSIM & 52.80 & 85.96 & 1.99$\times$ & 1.02$\times$ \\
eX2L-Soft Dice & 52.11 & 81.78 & 1.96$\times$ & 0.97$\times$ \\
\bottomrule
\end{tabular}
\end{table}

\section{Architecture Generalization Analysis} \label{appendix:e}

To evaluate whether eX2L's gains are tied to a specific backbone, we compared eX2L against ERM and GroupDRO on the \textit{Hard Many-to-Many Spawrious} benchmark using two additional architectures: a legacy backbone (AlexNet) and a modern high-capacity backbone (ConvNeXt-Base). All results use the best eX2L variant (MAE) and Uniform Group Sampling for eX2L and GroupDRO.

\begin{table}[ht]
\caption{Performance on AlexNet (Constrained) on \textit{Hard M2M Spawrious}}
\label{table:alexnet}
\centering
\small
\setlength{\tabcolsep}{4pt}
\begin{tabular}{@{}lcc@{}}
\toprule
\textbf{Algorithm} & \textbf{WGA} & \textbf{AA} \\
\midrule
eX2L-MAE & 3.17\% & 28.25\% \\
ERM       & 3.82\% & 28.20\% \\
GroupDRO  & 2.84\% & 28.05\% \\
\bottomrule
\end{tabular}
\end{table}

\begin{table}[ht]
\caption{Performance on ConvNeXt-Base (High-Capacity) on \textit{Hard M2M Spawrious}}
\label{table:convnext}
\centering
\small
\setlength{\tabcolsep}{4pt}
\begin{tabular}{@{}lcc@{}}
\toprule
\textbf{Algorithm} & \textbf{WGA} & \textbf{AA} \\
\midrule
\textbf{eX2L-MAE} & \textbf{50.44\%} & \textbf{73.41\%} \\
ERM                & 34.44\%          & 62.61\% \\
GroupDRO           & 47.66\%          & 71.01\% \\
\bottomrule
\end{tabular}
\end{table}

\paragraph{Architectural Bottleneck.} As per Table \ref{table:alexnet}, in the AlexNet regime, all algorithms converge to a near-zero WGA floor. This suggests the bottleneck is the backbone's representational capacity rather than the optimization strategy; the marginal differences are statistically negligible because AlexNet cannot encode group-invariant features effectively.

\paragraph{Explanation Fidelity and Scaling.} As per Table \ref{table:convnext}, with ConvNeXt-Base, eX2L substantially outperforms both ERM and GroupDRO ($+15.99\%$ WGA over ERM; $+2.78\%$ WGA over GroupDRO). This demonstrates that eX2L's efficacy is tied to the backbone's ability to produce spatially precise Grad-CAM maps. AlexNet's limited receptive field yields noisy, low-resolution heatmaps that undermine the similarity penalty, while ConvNeXt-Base provides the spatial granularity required to discriminate spurious from invariant regions. These results corroborate observations by \citet{angarano2024backbones} on the positive relationship between backbone capacity and domain generalization performance, and further show that eX2L exhibits superior scaling properties relative to ERM and GroupDRO, extracting greater gains from increased backbone capacity. The ResNet-50 results in the main paper thus evaluate eX2L at its high-capacity ceiling rather than its representational floor.

\section{Limitations and Future Work} \label{appendix:f}

\paragraph{Confounder Label Dependency.}
eX2L requires image-level confounder annotations at training time to supervise the auxiliary confounder classifier. This assumption is justified in high-stakes deployment settings such as medical imaging, autonomous driving, and algorithmic fairness auditing, where identifying confounders is already a regulatory or operational necessity, and the costs of annotation are outweighed by the risks of a biased model. Nevertheless, this requirement limits applicability in exploratory settings where nuisance attributes are unknown or costly to label. A natural extension is to replace the supervised confounder classifier with an unsupervised or self-supervised confounder discovery mechanism (e.g., clustering-based approaches, or the detection framework of \citet{dammu2023detecting}), enabling annotation-free debiasing. We leave this as an important direction for future work.

\paragraph{Scope and Generality.}
eX2L was primarily evaluated on binary and multi-class image classification benchmarks with known spurious backgrounds or attribute confounders. While the Spawrious Many-to-Many Hard results demonstrate robustness to compositional, multi-environment shifts, extension to multi-label classification, non-image modalities, and transformer-based architectures (e.g., Vision Transformers) remains an open research question. The architecture generalization analysis in Appendix~\ref{appendix:e} shows that eX2L's efficacy scales with backbone capacity, suggesting that higher-fidelity explanations available in modern architectures may further amplify performance gains. Extension to NLP settings (e.g., penalizing overlap between token attribution maps for class and confounder classifiers) is a promising direction given the widespread use of Grad-CAM analogues in transformer attention.

\paragraph{Comparison with Model-Debiasing Methods.}
eX2L is designed as a single-stage loss-based regularizer and was benchmarked against methods of the same class (ERM, IRM, CORAL, DANN, GroupDRO) following the WILDS and DomainBed evaluation protocols \citep{Koh2020WILDSAB, gulrajani2021in}. Two-stage model-debiasing methods such as Just Train Twice (JTT) \citep{liu2021justtraintwiceimproving}, Deep Feature Reweighting (DFR) \citep{kirichenko2023layerretrainingsufficientrobustness}, and Learning from Failure (LfF) \citep{nam2020learning}, which also require group annotations, operate under different assumptions (they discover spurious correlations implicitly from failure patterns on a biased model and then fine-tune the model afterwards) and require a separate reweighting or fine-tuning stage, making direct side-by-side comparison in a single-training-run protocol non-trivial. A unified benchmark comparing single-stage regularizers with two-stage debiasing methods under identical group annotation budgets would be a valuable contribution to the community and is a direction we encourage in future work.

\paragraph{Computational Overhead.}
Computing Grad-CAM heatmaps during training approximately doubles wall-clock time relative to ERM (see Appendix~\ref{appendix:d}). For very large-scale datasets or high-resolution inputs, this overhead may be prohibitive. Potential mitigations include gradient checkpointing, approximating Grad-CAM with cached activations, or using lighter attribution variants (e.g., GradCAM++) as drop-in replacements. The time-performance trade-off demonstrated across all benchmarks indicates that this cost is justified in difficult distribution shift settings where baseline methods fail.


\bibliography{egbib}
\end{document}